**Title:** Dendrites endow artificial neural networks with accurate, robust and parameter-efficient learning

**Authors:** Spyridon Chavlis[1], Panayiota Poirazi[1,*]

**Affiliations:** [1]Institute of Molecular Biology and Biotechnology, Foundation for Research and Technology-Hellas, Heraklion, Crete 70013, Greece; [*]corresponding author: poirazi@imbb.forth.gr

## Abstract

Artificial neural networks (ANNs) are at the core of most Deep learning (DL) algorithms that successfully tackle complex problems like image recognition, autonomous driving, and natural language processing. However, unlike biological brains who tackle similar problems in a very efficient manner, DL algorithms require a large number of trainable parameters, making them energy-intensive and prone to overfitting. Here, we show that a new ANN architecture that incorporates the structured connectivity and restricted sampling properties of biological dendrites counteracts these limitations. We find that dendritic ANNs are more robust to overfitting and outperform traditional ANNs on several image classification tasks while using significantly fewer trainable parameters. These advantages are likely the result of a different learning strategy, whereby most of the nodes in dendritic ANNs respond to multiple classes, unlike classical ANNs that strive for class-specificity. Our findings suggest that the incorporation of dendritic properties can make learning in ANNs more precise, resilient, and parameter-efficient and shed new light on how biological features can impact the learning strategies of ANNs.

## Introduction

The biological brain is remarkable in its ability to quickly and accurately process, store, and retrieve vast amounts of information while using minimal energy[1]. Artificial Intelligence (AI) systems, on the other hand, are notoriously energy hungry[2–4] and often fail on tasks where biological systems excel, such as continual and transfer learning[5–7]. The most widely used AI method is deep learning (DL)[8], which is applied in areas like computer vision[9] and natural language processing[10] and can even achieve superhuman performance in very specific tasks[11,12]. However, the number of trainable parameters needed to achieve such performance is large leading to generalization failures due to overfitting[13], as well as energy consumption levels that are not sustainable[14]. Moreover, unlike the brain, DL methods still fail to achieve high-performance accuracy under noisy settings[15,16] and tasks where information changes in a continuous manner[17]. This dichotomy between biological and artificial intelligence systems suggests that drawing inspiration from the brain may help enhance the efficiency of DL models, bringing them one step closer to emulating the biological way of information processing.

DL architectures rely heavily on multilayered artificial neural networks (ANNs) inspired by their biological counterparts. In these networks, artificial nodes are typically constructed as linearly weighted sums of their inputs followed by a nonlinearity, roughly imitating how the soma or axon of biological neurons integrates inputs[18], and learning occurs via changes in the connection strengths (weights) between these nodes[19]. In contrast, biological neurons are much more complex, consisting of a soma, an axon, and numerous dendrites that enable them to process thousands of synaptic inputs in parallel, in ways that differ extensively between cell types[20]. Although the somatic and axonal functionalities of biological neurons are well captured in artificial neurons, the dendritic computations are currently missing.

Biological dendrites, because of their ability to generate local regenerative events (dendritic spikes)[21,22], share a similar spiking profile as the neuronal soma. As a result, biological neurons can act as multi-layer ANNs[23–26], able to perform complex computations[27,28], such as logical operations[29,30], signal amplification and segregation[31,32], coincidence detection[33–36], multiplexing[37] and filtering of irrelevant or noisy stimuli[38,39]. Consequently, dendrites are thought to underlie complex brain functions, including perception[40,41], motor behavior[42,43], fear learning[44–46], and memory linking[47]. Moreover, dendrites can help achieve such functions in an efficient manner. For example, they enable learning with few plastic synapses[48], forming memories using small neuronal populations[24], and increasing storage capacity[49,50]. Given the high computational power of dendrites and the associated benefits in biological networks[27,51], the current design of artificial neurons seems outdated. Incorporation of dendritic properties would likely empower ANNs[52–54], fostering more effective, efficient, and resilient learning behaviors like those seen in biological networks.

The above proposition is supported by recent studies that have integrated dendritic structures and their properties into traditional ANNs[55–58], showing promising results on machine learning (ML) tasks[59–64]. For instance, adding active dendrites in ANNs was shown to enhance the network's ability to learn continually[63], while including a specific dendritic nonlinearity improved performance in a multitask learning scenario[65]. However, to achieve improved performance, these studies have either sacrificed biological plausibility[64], used a very large number of trainable parameters[63], or were applied to very simple tasks[58].

Here, we propose a bio-realistic dendritic architecture that aims to improve learning in ANNs trained with the backpropagation algorithm. In the proposed architecture, inputs are fed into the dendritic layer, which is, in turn, connected to the somatic layer in a sparse and highly structured manner (**Figure 1**). Moreover, input sampling is inspired by the receptive fields of neurons in the visual cortex[66,67] and amounts to sampling a restricted part of the input, similar to a specific form of convolutional networks, the so-called locally-connected networks[68], as opposed to the entire image that is typically done in ANNs. By incorporating dendritic structural and sampling features, the new dendritic ANN models match or outperform traditional ANNs on several image classification tasks and counteract overfitting, while using orders of magnitude fewer trainable parameters. These improvements are likely due to a more extensive utilization of trainable weights and a different learning strategy used by dendritic versus traditional ANNs. Overall, our findings suggest that dendrites can augment the computational efficiency of ANNs without sacrificing their performance accuracy, opening new avenues for developing bio-inspired ML systems that inherit some of the major advantages of biological brains.

## Results

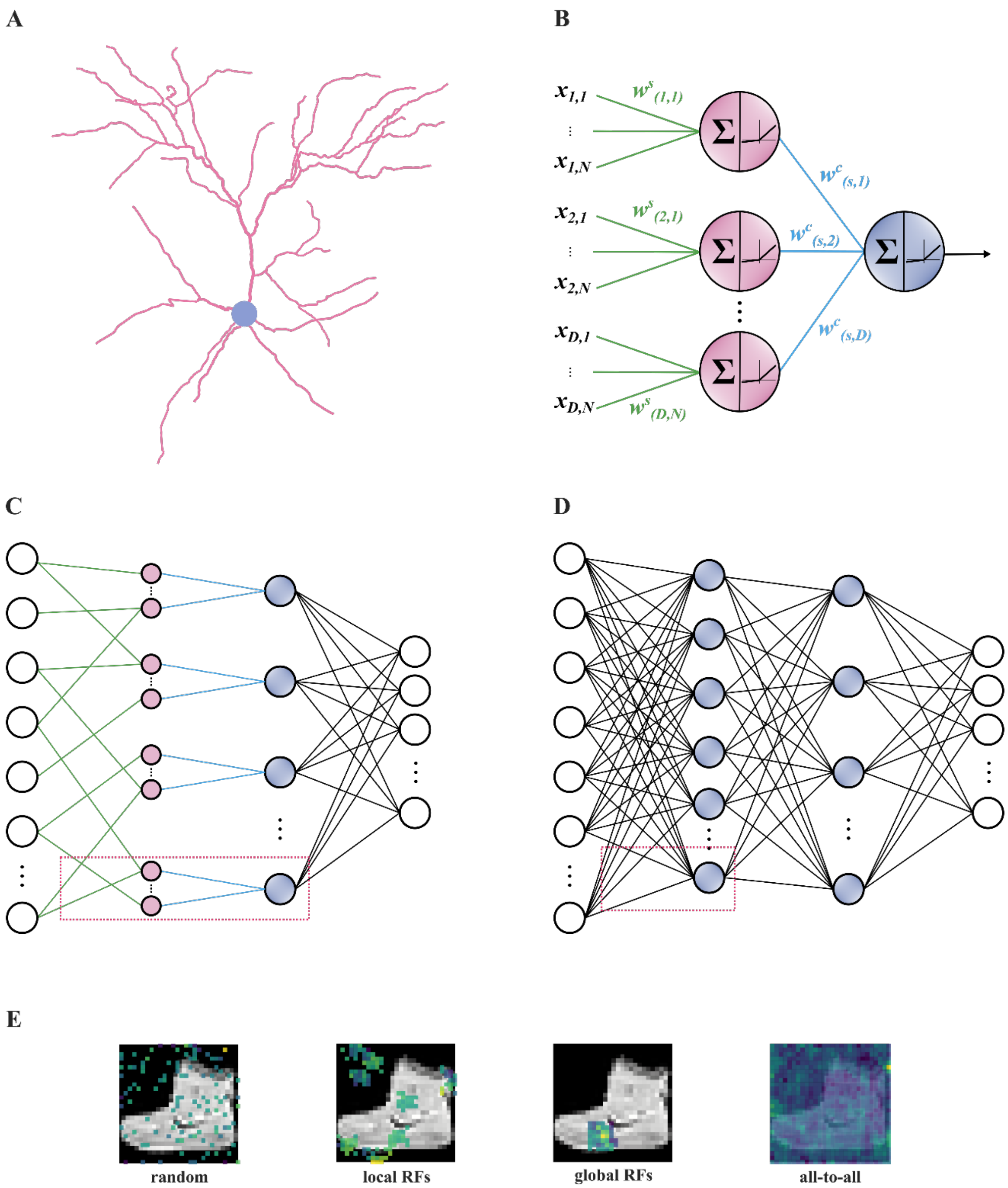

**Figure 1. Schematic representation of the dendritic ANN (dANN) compared to a classical vanilla ANN (vANN). A.** Example of a layer 2/3 pyramidal cell of the mouse primary visual cortex (dendrites: pink; soma: grey) that served as inspiration for the artificial dendritic neuron in **B**. The morphology was adopted from Park et al.[69]. **B.** The dendritic neuron model consists of a somatic node (blue) connected to several dendritic nodes (pink). All nodes have a nonlinear activation function. Each dendrite is connected to the soma with a (cable) weight, $w^{c}_{(d,s)}$, where *d* and *s* denote the dendrite and soma indices, respectively. Inputs are connected to dendrites with (synaptic) weights, $w^{s}_{(d,n)}$, where *d* and *n* are indices of the dendrites and input nodes, respectively. $d \in \{1, D\}, n \in \{1, N\}$, *N* denotes the number of synapses each dendrite receives, and *D* the number of dendrites per soma *s*. **C.** The dendritic ANN architecture. The input is fed to the dendritic layer (pink nodes), passes a nonlinearity, and then reaches the soma (blue nodes), passing through another nonlinearity. Dendrites are connected solely to a single soma, creating a sparsely connected network. **D.** Typical fully connected ANN with two hidden layers. Nodes are point neurons (blue) consisting only of a soma. **E.** Illustration of the different strategies used to sample the input space: random sampling (R), local receptive fields (LRF), global receptive fields (GRF) and fully connected (F) sampling of input features. Examples correspond to the synaptic weights of all nodes that are connected to the first unit in the second layer. The colormap denotes the magnitude of each weight.

To explore the role of dendritic properties in efficient learning, we developed a dendritic ANN (dANN) model with structured connectivity that loosely mimics the morphology of biological neurons (**Figure 1A**). In this model, each dendrite acts as a typical point neuron: it linearly sums its weighted inputs (synapses) and passes the sum through a nonlinearity. The dendritic activations are subsequently multiplied by the cable weights and summed at the soma before going through a second nonlinearity (**Figure 1B**). To train the model using ML platforms (e.g., TensorFlow, PyTorch, Jax), we implemented it as a traditional ANN with two sparsely connected hidden layers, representing the dendritic and somatic units, respectively, and a fully connected output layer (**Figure 1C**). For comparison purposes, we also implemented a fully connected, vanilla ANN (vANN) with the same number of layers (**Figure 1D**).

In addition to their structured connectivity, the dendrites of biological neurons typically receive only partial information from a visual scene[70]. To investigate the contribution of this property, we implemented four types of input sampling for the dendritic ANN model: a) random sampling of input features (R), b) local receptive fields (LRF) where each dendrite samples from a spatially restricted part of the image, c) global receptive fields (GRF) where all dendrites belonging to a soma sample from the same spatially restricted part of the image, and d) an all-to-all, fully connected type of sampling (F), which is also used by the vANN (**Figure 1E**). We then tested the learning capabilities of our models on various image classification tasks (**Supplementary Figure 1**) using the same (default) hyperparameters, optimization algorithm, and loss function (see *Methods*). To ensure fair comparisons, we tested equivalent network architectures for all models, i.e., consisting of the same number of nodes in each hidden layer.

**Bio-inspired dendritic ANNs are more accurate, robust, and efficient than vanilla ANNs on image classification.**

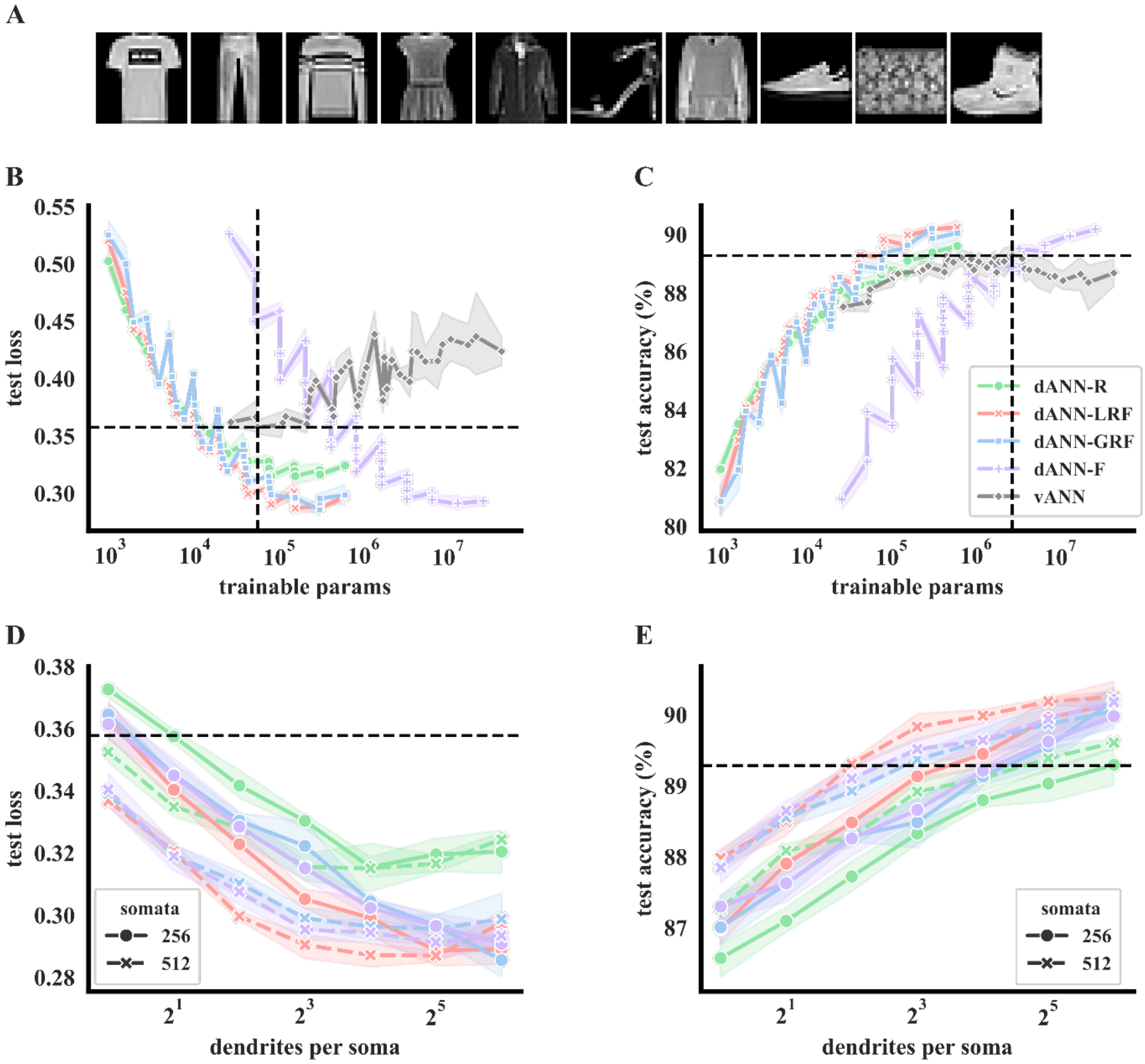

**Figure 2. Dendritic features improve learning on Fashion MNIST classification. A.** The Fashion MNIST dataset consists of 28x28 grayscale images of 10 categories. **B.** Average test loss as a function of the trainable parameters of the five models used: A dendritic ANN with random inputs (dANN-R, green), a dANN with LRFs (red), a dANN with GRFs (blue), a dANN with all-to-all inputs (dANN-F, orange), and the vANN with all-to-all inputs (grey). Horizontal and vertical dashed lines denote the minimum test loss of the vANN and its trainable parameters, respectively. The *x*-axis is shown in a logarithmic scale (log10). **C.** Similar to **B**, but depicting the test accuracy instead of the loss. **D.** Test loss as a function of the number of dendrites per somatic node for the four dANN models. The linestyle (solid and dashed) represents

different somatic numbers. The dashed horizontal line represents the minimum test loss of the vANN (512-256 size of its hidden layers, respectively). The *x*-axis is shown in a logarithmic scale (log2). **E.** Similar to **D**, but showing the test accuracy instead of the loss. The dashed horizontal line represents the maximum test accuracy of the vANN (2048-512 size of its hidden layers, respectively). For all panels, shades represent three standard deviations across *N=5* initializations for each model.

We first tested the learning capabilities of all models against the Fashion MNIST (FMNIST) dataset (**Figure 2A**). We found that dANNs with restricted input sampling achieve better learning and combat overfitting much more effectively than vANN models for both size-matched and larger vANN architectures (in terms of trainable parameters). This is evidenced by a consistently lower test loss for all dANN models - except the dANN-F - compared to vANNs of the same number of trainable parameters (**Figure 2B**). Importantly, the vANN models exhibit overfitting as the model size increases (**Figure 2B** and **Supplementary Figure 2**), while this does not occur for dANNs (for the model sizes tested), suggesting that dendrites may serve as natural regularizers[71]. Indeed, reductions in overfitting are also evident in vANNs when using various regularization methods or hyperparameter tuning such as dropout (**Supplementary Figure 3**), different learning rates (**Supplementary Figure 4**) or an early stopping criterion (**Supplementary Figure 5**). These overfitting reductions, however, are not as large as those seen in dANNs, suggesting that dendrites provide a more robust regularization effect.

In addition to overfitting benefits, dANNs with restricted input sampling match the best performance of vANNs via the use of much fewer trainable parameters (**Figure 2C**), suggesting that dendritic features render ANNs more efficient. Among the four dANN configurations tested, the dANN with local receptive fields (dANN-LRFs) is the most efficient: it reaches maximum accuracy and minimum loss with over one order of magnitude fewer trainable parameters than the vANN. The dANN models with random (dANN-R) and global receptive field (dANN-GRF) sampling are slightly worse but significantly more efficient than the vANN, while the dANN with all-to-all input sampling (dANN-F) shows reduced overfitting but no efficiency gains with respect to accuracy. Overall, these findings suggests that both dendritic features, i.e., the structured dendritic connectivity and the restricted input sampling, contribute to the efficiency gains of dANN models.

Finally, we found that, as expected, learning in dANN models improves with network size (lower loss: **Figure 2D**, better accuracy: **Figure 2E**). More importantly, unlike other bio-inspired architectures[72], dANN models appear to scale well with increasing depth (**Supplementary Figure 6**), revealing their potential for use in deeper architectures.

To substantiate our results on the beneficial role of dendritic features, we tested the dANN models on five additional benchmark datasets (**Figure 3, Table 1**, *see* §*Datasets*). As with FMNIST, we found that the best dANN models consistently outperformed - albeit slightly- the best vANN in terms of both accuracy and loss (**Table 1** and **Supplementary Table 1**). Moreover, similarly to FMNIST, we found that dANNs with restricted input sampling (i.e., R, LRF, GRF) are much more efficient than vANN for all datasets. Specifically, they can match the accuracy (**Figure 3A**) and loss (**Figure 3B**) of the best vANN using 1-3 orders of magnitude fewer trainable parameters. It is worth noting that for more difficult tasks, like the CIFAR10 dataset, the difference in both the number of trainable parameters (**Figure 3A-B)** and the best accuracy (**Table 1**) between these dANNs and vANNs is more prominent. Finally, the all-to-all input sampling enables dANNs to achieve slightly higher accuracy than vANNs on certain datasets (**Table 1**), albeit with a greater number of trainable parameters (**Figure 3A-B**, purple bars), thus diminishing the efficiency gains of dANNs.

To quantify the efficiency differences between dANNs and vANNs, we formulated the efficiency score metrics, which normalize the best accuracy (**Figure 3C**) and the corresponding loss (**Figure 3D**) that a given model can achieve with the number of trainable parameters used multiplied by the number of epochs needed to reach minimum validation loss (see *Methods*). These metrics consider both the size of a network (trainable parameters) and a proxy of its convergence speed (training epochs). We found that dANN models with restricted input sampling exhibit higher efficiency than vANNs across all datasets. This is not the case for the dANN with all-to-all input sampling, which seems to be more efficient than vANNs in terms of minimum loss but less efficient in terms of maximum accuracy (**Figure 3C, D**). This is because the dANN-F model has a very large number of parameters and needs a large number of epochs for training (**Supplementary Figure 7**), yet does not suffer from overfitting, thus achieving a smaller loss for similar accuracy in the majority of the datasets tested.

**Table 1.** Top test accuracy scores obtained by each model on five benchmark datasets across various configurations and their corresponding test loss. Performance accuracy and loss are listed as mean ± standard deviation over *N=5* initializations for each model.

| | **MODEL PERFORMANCE – TOP ACCURACY** | | | | |
|---|---|---|---|---|---|
| **Models** | **MNIST** | **FMNIST** | **KMNIST** | **EMNIST** | **CIFAR10** |
| | **Test accuracy (%)** | | | | |
| dANN-R | 98.090±0.0583 | 89.612±0.0870 | 91.076±0.0811 | 82.745±0.2568 | 52.458±0.4347 |
| dANN-LRF | 98.466±0.1058 | **90.256±0.2237** | **91.928±0.0643** | 83.166±0.0893 | 56.966±0.9796 |
| dANN-GRF | **98.576±0.0809** | 90.182±0.2164 | 90.046±0.2915 | 83.779±0.2064 | 56.998±0.4875 |
| dANN-F | 98.466±0.0862 | 90.184±0.2203 | 90.61±0.6927 | **84.121±0.1880** | **60.12±0.3906** |
| vANN | 98.034±0.2742 | 89.288±0.3654 | 91.552±0.6629 | 83.381±0.2681 | 49.082±1.2092 |
| | **Test loss** | | | | |
| dANN-R | 0.0644±0.0013 | 0.3245±0.0028 | 0.5068±0.0045 | 1.0591±0.0225 | 1.4612±0.0172 |
| dANN-LRF | 0.0483±0.0018 | 0.2975±0.0042 | **0.4374±0.0037** | 0.6105±0.0048 | **1.2684±0.0230** |
| dANN-GRF | **0.0471±0.0022** | 0.2947±0.0052 | 0.5757±0.0335 | 0.5605±0.0077 | 1.3398±0.0217 |
| dANN-F | 0.0503±0.0023 | **0.2935±0.0043** | 0.4789±0.0378 | **0.5390±0.0060** | 1.2811±0.0247 |
| vANN | 0.0967±0.0116 | 0.4040±0.0066 | 0.8246±0.0791 | 2.0860±0.0476 | 2.0455±0.1375 |

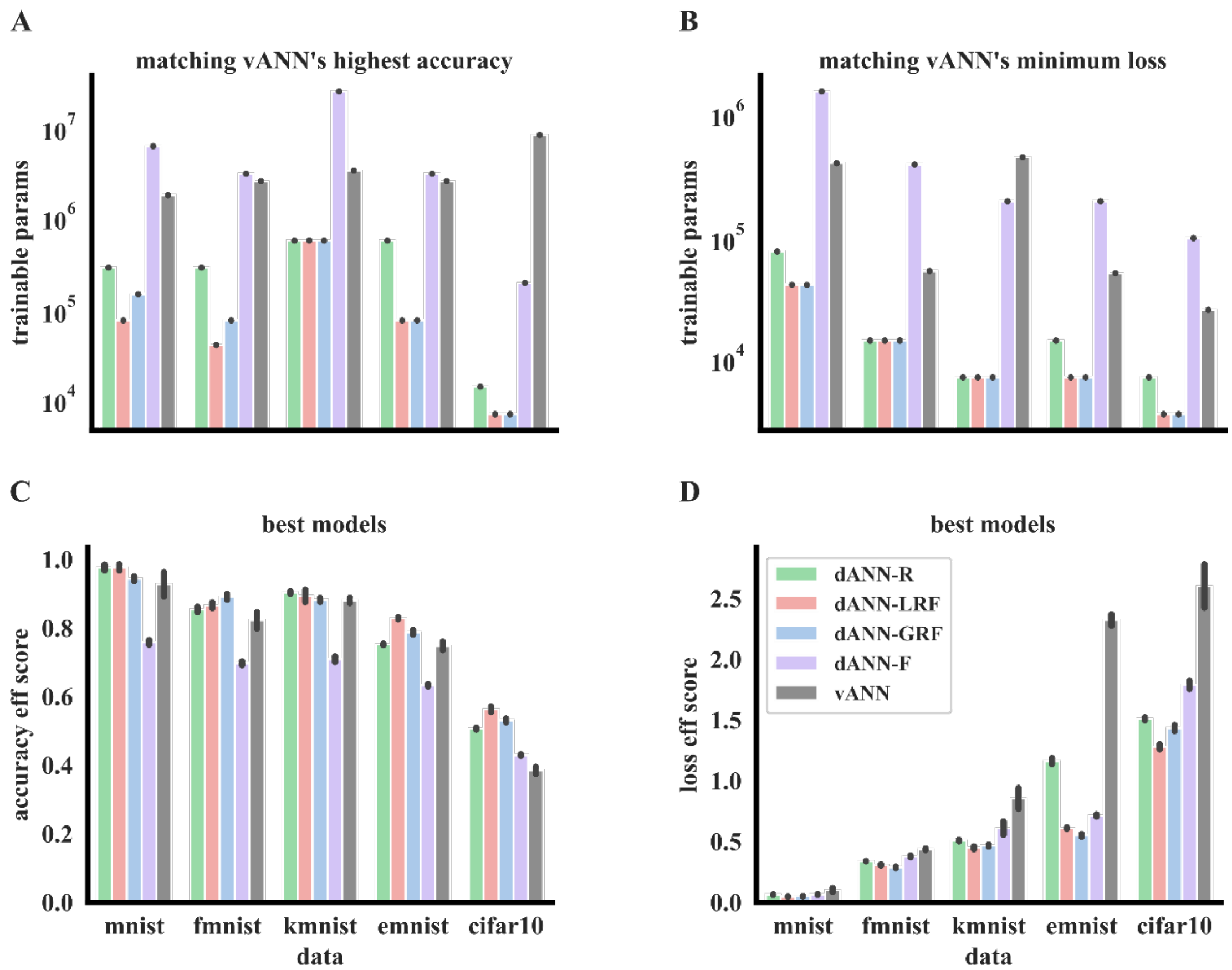

**Figure 3. Dendrites improve performance across various benchmark datasets. A.** Number of trainable parameters that each model needs, dANN-R (green), dANN-LRF (red), dANN-GRF (blue), and dANN-F (orange) to match the highest test accuracy of the respective vANN (grey). **B.** The same as in **B**, but showing the number of trainable parameters required to match the minimum test loss of the vANN (grey). **C.** Accuracy efficiency score for all models and all datasets tested. Test accuracy is normalized with the logarithm of trainable parameters. The score is bounded in [0, 1]. **D.** Same as in **C**, but showing the loss efficiency score. Again, we normalized the test score with the logarithm of the trainable parameters times the number of epochs needed to achieve minimum validation loss. The score is bounded in [0, ∞). In all barplots the errorbars represent three standard deviations across $N=5$ initializations for each model.

The above findings illustrate that the two dendritic features implemented here, i.e., the structured connectivity and the restricted input sampling, provide important efficiency gains on image classification compared to classical vANNs. To dissect the relative contributions of these two features, we compared the dANN models to densely connected vANN models (**Supplementary Figure 8** and **Supplementary Figure 9**) and sparsely connected sANN models (**Supplementary Figure 10** and **Supplementary Figure 11**), furnished with the four types of input sampling. We found that while restricted sampling improves the performance efficiency of vANNs, dANNs still outperform these improved models (**Supplementary Figure 8**) suggesting that restricted sampling alone cannot account for the improved efficiency of dANN models.

To assess the contribution of structured dendritic connectivity, we next compared dANNs to randomly connected, sparse ANN models (sANNs). Sparse neural networks were previously shown to exhibit improved performance[73] and since dANNs are a specific subset of sANNs, it is likely that efficiency gains stem primarily from internal sparsity. We found that the bio-inspired dANNs, namely those with structured restricted input sampling (LRF and GRF), consistently outperform sANNs in terms of efficiency gains. When comparing dANNs and sANNs with matched input sampling types -whereby the only difference is the structured vs. random internal connectivity-, differences in efficiency gains are reduced. However, dANNs still exhibit higher efficiency gains (**Supplementary Figure 10**) suggesting that a structured, tree-like internal sparsity provides additional efficiency benefits.

Overall, these experiments confirm that both dendritic features, namely the structured connectivity and the restricted input sampling, contribute significantly to improving the performance accuracy of dANNs, their efficiency and their resilience to overfitting. Given these findings, in the following sections we focus our analysis on the bio-inspired dANN models (i.e., those with restricted sampling: dANN-R, dANN-LRF, dANN-GRF) and fully connected vANN models.

## Bio-inspired dANNs employ a distinct learning strategy and fully exploit their available resources

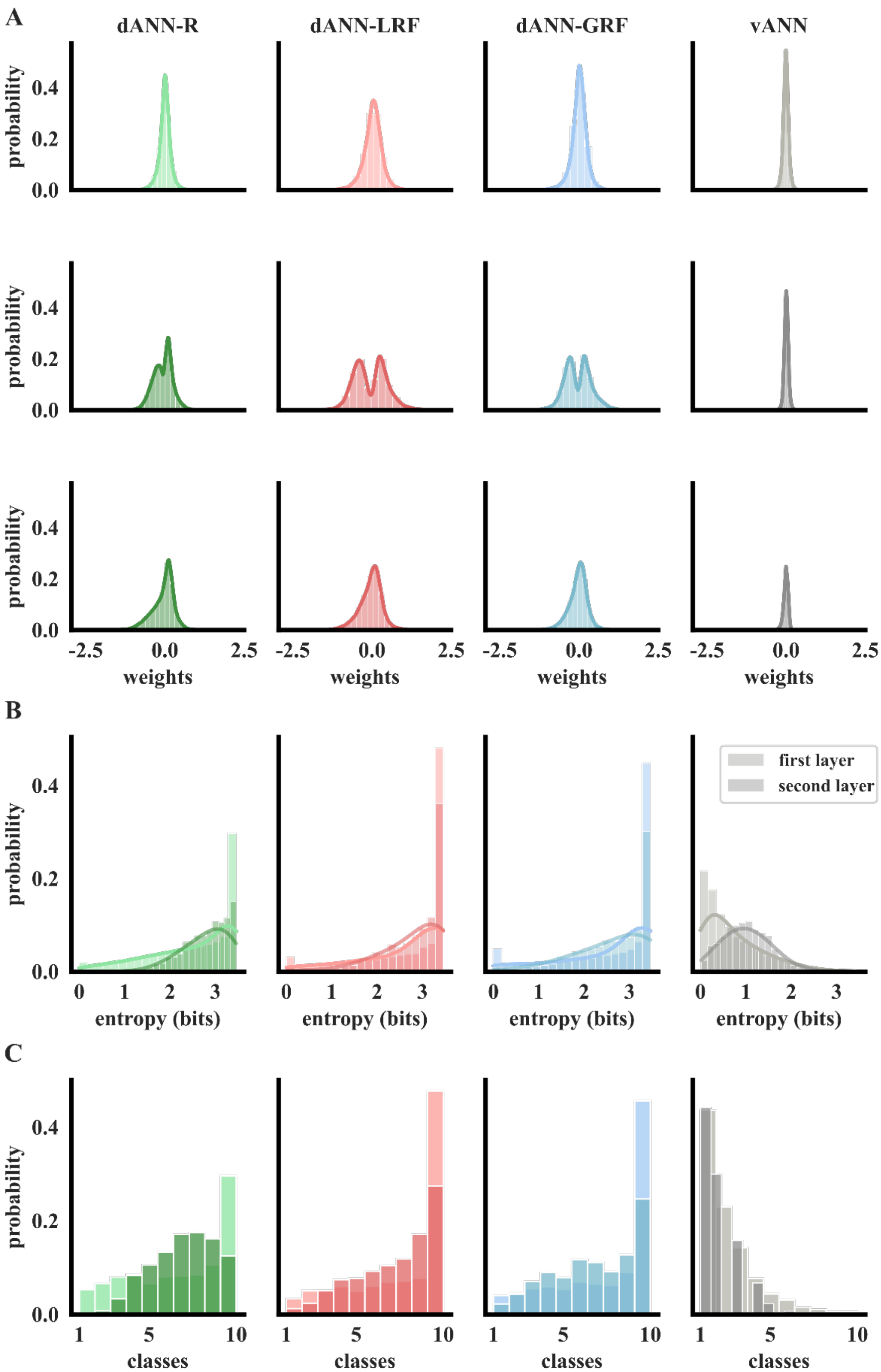

**Figure 4. Bio-inspired dANN models fully exploit their available resources and solve the task using a different learning strategy. A.** Weight probability density functions after training for dANN-R, dANN-GRF, dANN-LRF, and vANN. The density functions are built by concatenating all weights across *N=5* initializations for each model. First hidden layer (top row), second hidden layer (middle row), and output layer (bottom row) weights are shown. Both *x* and *y* axes are shared across all subplots for visual comparison among the density plots. **Supplementary Table 2** contains the kurtosis, skewness, and range of all KDE plots. **B.** Probability density function of the entropy (bits) for the first (normal color) and second (shaded color) hidden layer, respectively. Entropies are calculated using the activations of each layer for all test images of FMNIST (see *Methods*). Silent nodes have been excluded from the visualization. Higher values signify mixed selectivity, whereas low values indicate class specificity. **C.** Probability density functions of selectivity for both layers (different color shades) and all models (columns). For all histograms, the bins are equal to the number of classes, i.e., for the FMNIST dataset.

To better understand why dANN models with restricted input sampling outperform vANNs, we analyzed their weight distributions post-learning of the Fashion MNIST dataset. We found a broader distribution, i.e., larger range of values, of synaptic (layer 1) weights for dANNs compared to vANN (**Supplementary Table 2** and **Figure 4A**, top row) and a bimodal distribution of dANN cable (layer 2) weights, all centered around zero (**Supplementary Table 2** and **Figure 4A**, middle row). For the dANN-LRF model, in particular, there were very few cable weights close to zero, indicating that the model effectively utilizes all trainable parameters of this layer. In contrast, in the vANN model, weights follow a Gaussian-like distribution centered around zero, suggesting that many weights are not as effectively utilized. Finally, the distribution of the output layer weights in dANNs is broader than in the vANN model (**Supplementary Table 2** and **Figure 4A**, bottom row). These observations suggest that bio-inspired dANNs fully exploit their trainable parameters, especially their cable (second layer or dendrosomatic) weights, compared to the second hidden layer of the vANN.

To delineate how the nodes of the different models contribute to a decision, we looked into their selectivity. First, we calculated the information entropy, which measures how class-specific a node is. High entropy values indicate mixed selectivity, whereby the node is active for more than one class, while low values indicate class specificity. We found opposite entropy distributions between the bio-inspired dANNs and the vANN. This means that the dANN models primarily have mixed-selective nodes in both hidden layers, while vANNs primarily have class-specific nodes. This difference was even more pronounced for dANNs with global or local RFs (**Figure 4B**).

To assess whether the observed differences in entropy map onto node specificity, we formulated the selectivity index, which counts how many classes a given node responds to. Specifically, if a node is active (activation greater than zero for a given image) for more than 400 images of a specific class, corresponding roughly to 40% of testing images, its selectivity index for that class is set to one. As with entropy distributions, we found that in bio-inspired dANNs, both layers consist primarily of mixed-selective nodes, while the vANN contains primarily class-specific nodes (**Figure 4C**). These observations suggest that dANN and vANN models employ different strategies to solve the same classification task.

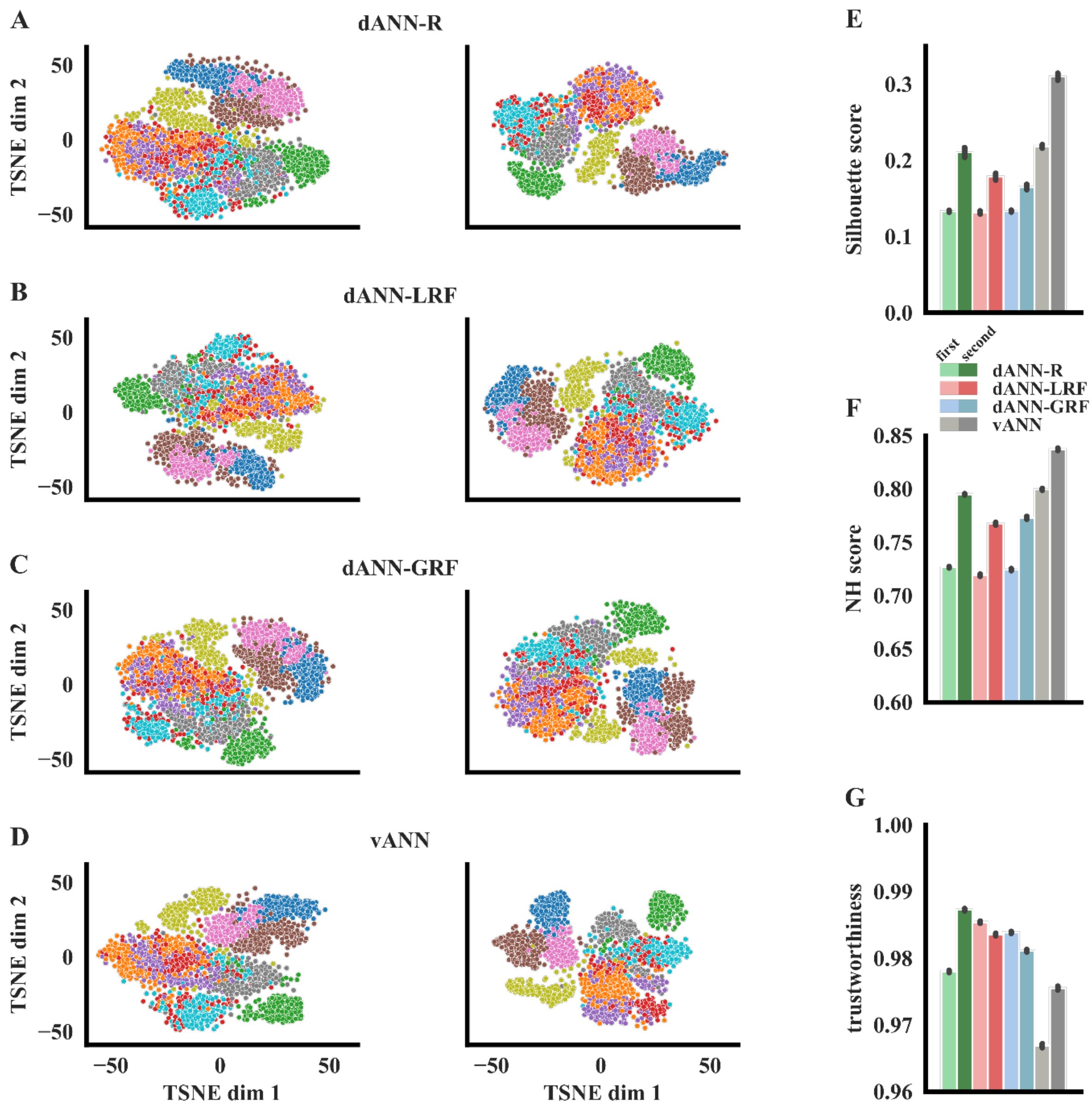

**Figure 5. Learned representations. A-D.** TSNE projections of the activations for the first (left column) and the second (right column) hidden layers corresponding to the three dANN and the vANN models. Different colors denote the image categories of the FMNIST dataset. While the figure shows the results of one run, the representations are consistent across 10 runs of the TSNE algorithm (data not shown). **E.** Silhouette scores of the representations (2-way ANOVA: model F(3,32)=1598.31, p<10-3, layer F(1,32)=2130.39, p<10-3, model x layer F(3,32)=105.20, p<10-3). **F.** Neighborhood scores of the representations, calculated using 11 neighbors (2-way ANOVA: model F(3,32)=8624.78, p<10-3, layer F(1, 32)=18512.42, p<10-3, model x layer F(3, 32)=299.51, p<10-3). **G.** Trustworthiness of the representations, calculated using 11 neighbors (2-way ANOVA: model F(3,32)=6187.66, p<10-3, layer F(1,32)=1856.84, p<10-3, model x layer F(3, 32)=1777.98,

p<10-3). In all barplots the errorbars represent three standard deviations across *N=5* initializations for each model and 10 runs of the TSNE algorithm per initialization. Stars denote significance with unpaired t-test (two-tailed) with Bonferroni's correction.

To complete our interpretability analysis, we visualized the hidden representations of the compared dANN and vANN models post-learning. The goal was to assess the amount of high-level information that is extracted by the first and second hidden layers across models (i.e., dendritic and somatic layers for dANNs, respectively). We applied the T-distributed stochastic neighbor embedding (TSNE), an algorithm that reduces the dimensionality and allows visualization of high-dimensional data[74]. By visual inspection, we observed a change in the representation between the dendritic and somatic layers of dANN models, similarly to representations of vANN between its two hidden layers (**Figure 5A-D**). We quantified the separability of the representations using the silhouette and the neighborhood (NH) scores, which measure the global and local degree of separability, respectively (see *Methods*). In the three dANN models, global and local separability was increased from the dendritic to the somatic layer (**Figure 5E-F**), something that we also observed in the hidden layers of the vANN. This means that the discriminatory power of both dANNs and vANN increases across layers in a similar way. This is in line with the findings of **Figure 4,** whereby the vANN is shown to have higher class-specificity than the dANNs in the first layer, and thus higher separability scores. Importantly, our results regarding the properties of the representations in low-dimensional space reflect the properties of the high-dimensional data as shown by their high trustworthiness scores (**Figure 5G**). The latter measures the extent to which the local structure of the data is retained after projection to the lower-dimensional space. Values close to 1 indicate high reliability. **Figure 5G** suggests that the three dANNs do a better job in retaining the structure of the original data in their representations, as measured by TSNE, compared to vANNs. This is probably a result of the different strategy implemented by these networks.

Overall, our interpretability analysis reveals that dANNs with restricted input sampling use a different strategy than the vANN model to achieve accurate, robust, and efficient image classification: rather than becoming class-specific early on like the vANN, the dANN models exhibit mixed-selectivity in both layers. This strategy may underlie their ability to create trustworthy representations of the input data, and achieve high performance accuracy and reduced overfitting, while using significantly fewer yet fully utilized trainable parameters.

## Dendritic benefits are more pronounced as the task difficulty increases

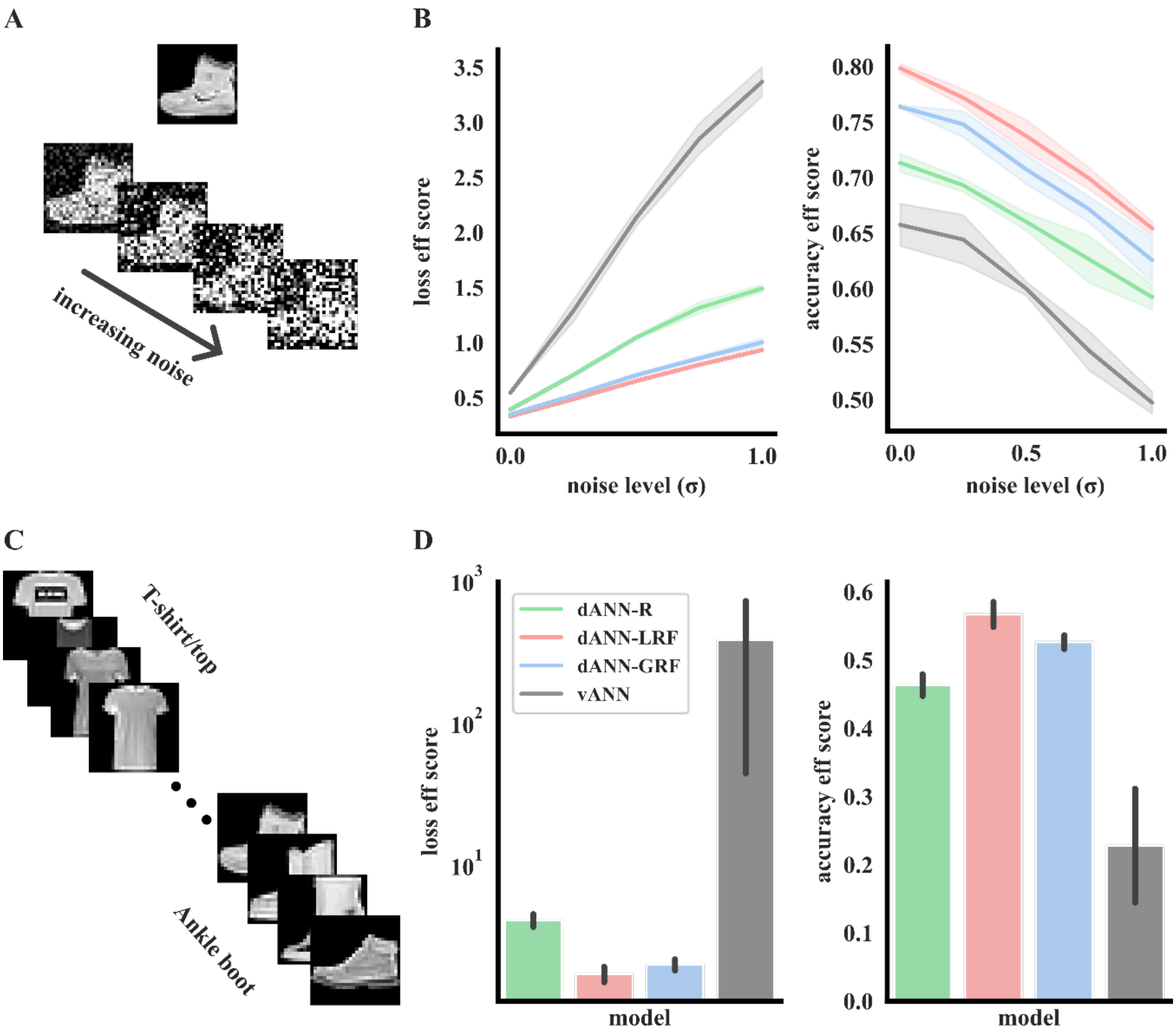

**Figure 6. Bio-inspired dANNs are more accurate and efficient than vANNs when inputs are noisy or presented in a sequential manner. A.** An example of one FMNIST image with variable Gaussian noise. Sigma (σ) is the standard deviation of the Gaussian noise. **B.** Testing loss (*left*) and accuracy (*right*) efficiency scores for all models and noise levels. Shades represent three standard deviations across *N=5* network initializations for each model. **C.** The sequential learning task. **D.** As in **B**, but showing the loss (left) and accuracy (right) efficiency scores for the sequential task. Errorbars denote three standard deviations across *N=5* initializations for each model. See **Table 2** and **Supplementary Table 3** for the accuracy and loss values.

Our image discrimination results suggest that the difference between bio-inspired dANN and vANN models may be larger for more difficult tasks/datasets (see results for CIFAR10 in **Table 1**). To test this hypothesis, we constructed learning scenarios that are known to be challenging for ANN models.

First, we added Gaussian noise (with a variable σ and zero mean) to all images in the FMNIST dataset, thus creating new datasets of increasing classification difficulty (**Figure 6A**). We then selected the best vANN and the corresponding dANNs that matched its performance accuracy on FMNIST (from **Figure 2** and **Figure 3**) and tested their performance on the noisy datasets. We found that, while the performance of all models declined with increasing noise levels, dANNs with restricted input sampling demonstrated higher efficiency and resilience. This is evident by a slower increase rate for the loss and a slower drop rate for the accuracy efficiency scores, respectively, compared to vANNs (**Figure 6B**). In all cases, the best-performing dANN was the one with local RFs (dANN-LRF).

To confirm the advantages of dANN models on challenging tasks, we constructed a second learning scenario that remains challenging for traditional ANNs. In this task, models were fed with batches of inputs belonging to the same class in a sequential manner (**Figure 6C**). This process, which was repeated 50 times (epochs), results in models receiving information only from images of a single class during gradient calculation. As with the noisy task, the three dANN models were more accurate (**Table 2**), less variable across different initializations, and much more efficient than the vANN, as evidenced by their loss and accuracy efficiency scores (**Figure 6D**). The best-performing dANN was again the one with local RFs (dANN-LRF). Overall, these findings suggest that incorporation of dendritic features in ANNs may result in even greater robustness, accuracy and efficiency gains when the task difficulty is increased.

**Table 2.** Test accuracy obtained by each model on the sequential learning task, and their corresponding test loss, using the FMNIST dataset. Test accuracy and loss are listed as mean ± standard deviation across *N=5* initializations for each model.

| | **MODEL PERFORMANCE – SEQUENTIAL LEARNING** | |
|---|---|---|
| **Models** | **Test accuracy (%)** | **Test loss** |
| dANN | 56.834±1.6748 | 3.3947±0.3063 |
| dANN-LRF | **59.658±1.5005** | **1.6532±0.1783** |
| dANN-GRF | 59.348±1.3412 | 1.8007±0.1537 |
| vANN | 28.458±9.4994 | 313.466±240.8529 |

## Discussion

Bio-inspired machine learning is one of the most dynamic branches of AI. Biological dendrites and their learning rules, are among the top candidates being explored, already showing highly promising results in artificial neural networks[55,56,75,76]. Recent studies in dendritic networks focus on their potential to tackle difficult problems such as continual[60,63] and multitask learning[65] and propose solutions to the credit assignment problem without backpropagation[56,57]. However, the processing power and efficiency of biological networks, largely endowed by their dendrites, are still far from being matched by respective ANNs. Towards this goal, we focused on the structured connectivity and restricted sampling characteristics of dendrites, two prominent features that are conserved across brain regions and species[77,78], suggesting that their role in information processing is likely to be very important.

We constructed a set of dendritic ANNs that leverage the structured connectivity and restricted sampling features of biological dendrites to enhance learning. dANNs are constructed as typical ANNs with two hidden layers, in which the first (dendritic) layer is connected in a sparse and structured manner to the second (somatic) layer so that it resembles the structured connectivity of biological dendrites to their respective somata. Input sampling was inspired by the receptive fields of neurons in the visual cortex[67,78], whereby dendrites (and neurons) sample only a restricted part of the visual space. We compared our models to vanilla ANNs across numerous image classification tasks and found that they are superior in performance accuracy, degree of overfitting, robustness to noise, and sequential learning. Compared to vANN, these benefits are achieved with orders of magnitude fewer trainable parameters, making dANNs more efficient and effective. As there is growing concern that the demand for computational resources to develop and apply AI models could lead to a significant increase in the electricity consumption of data centers worldwide[79], dANNs are especially valuable for edge computing and other energy-constrained scenarios. Even though we focused on simple ANNs that form the foundation of nearly all DL architectures, it is important to note that our study also offers a framework for integrating dendritic features into various models, such as convolutional neural networks, transformers, and others[80]. This involves replacing the fully connected layers commonly found in deep learning architectures with our dANN model.

To dissect the relative contributions of structured connectivity vs. restricted input sampling, we also compared dANN models to fully connected ANNs and sparse (randomly-connected) ANNs furnished with the input sampling types that were used in the dendritic models. We found that both features are important for achieving the overfitting and accuracy efficiency gains and that their relative contribution varies with the dataset and task used.

Of note, while some of the sparse networks approximate the performance of dANNs, the space of all sparse networks of a given size is vast, and finding those networks that work best is non-trivial. This work demonstrates that a biologically-inspired structured architecture (namely the dendritic one) can result in improved and more efficient performance. This finding is unexpected, as the bio-inspired architecture -including tree-like hidden layer connectivity and sampling with receptive fields- is just one of the too many possible sparse architectures. Knowing that nature has identified a network architecture that is more efficient than classical neuronal network architectures is important for designing efficient systems, without the need for extensive and expensive grid searches to identify such architectures.

Throughout these comparisons, the dANN model with LRF subsampling emerged as the most efficient configuration from a computational perspective. LRF subsampling assumes that individual dendrites of a given soma preferentially receive inputs from focused regions of a visual scene and that these regions are not necessarily close in visual space (**Figure 1**). In the visual cortex, this could amount to receiving clustered inputs from presynaptic neurons that share similar feature selectivity, allowing for selective and distributed information processing within the dendritic tree. Several studies in both the cortex and the hippocampus provide evidence that dendrites can sample from the same feature space and receive inputs with correlated tuning properties[40,81–84]. Additionally, dendrites can exhibit local, branch-specific integration and plasticity[85]. This feature-based input organization is further supported by observations of branch-specific dendritic depolarizations and the generation of local dendritic spikes in response to specific input combinations[29,86]. This structured organization of inputs has been suggested to confer important computational advantages for learning and memory storage[49,87–89].

Our interpretability analysis revealed that dANN models -unlike vANN models- form primarily mixed-selective nodes. This finding aligns with experimental evidence of mixed-selective neurons in various cortical regions of the mouse and primate brains[90–93]. Such neurons can encode multiple task-relevant features simultaneously, a property believed to be important for flexible decision-making and information processing, especially in higher cortical regions such as the prefrontal cortex. The ability to form and utilize mixed-selective nodes suggests that dANNs may enable more efficient and adaptable information processing, similar to biological neural networks. Further research into the similarities and differences between mixed-selective nodes in ANNs and mixed-selectivity neurons in the brain could provide valuable insights into the principles underlying intelligent behavior.

Beyond the findings reported here, other studies have also adopted dendritic properties in ANNs and DL models. These studies are different and complementary to ours in several ways. For example, one approach implemented dendrites as max-pooling or average-pooling layers[63,64,94], two methods that are extensively used in DL models[95]. Other approaches were more abstract, modeling dendrites as a multiplicative component and/or using all-to-all connectivity from the input layer onto dendrites[62,96–98]. Some methods use specific dendritic components, such as normalization of the weights or specific dendritic spikes, but their connectivity matrices become sparse using, for example, evolutionary algorithms, or they contain fully connected layers[65,99,100]. Lastly, dendritic models incorporating local learning rules have been used to study how the brain solves the credit assignment problem, showing promising learning capabilities but are not easily applicable to large ML applications[55,56,75].

Inspired by Jones and Kording[58], our modeling approach considers dendrites as an additional layer that provides weighted inputs to the somatic nodes. This means that dANN can be viewed as a sparse ANN from an ML perspective. In ANNs, sparsity can be achieved by pruning after training[101,102], using evolutionary algorithms[103,104], specific regularization[105,106] during training, or iterative methods applied before training[107,108]. Similar to the latter approaches, in dANNs, the connectivity sparsity is handcrafted from the start, creating a fixed architecture that is significantly smaller than typical vANNs undergoing a pruning process. This makes training faster and more efficient, as no pruning is required. Furthermore, inputs to dendrites are not randomly allocated but can be constructed based on RFs, setting our model apart from traditional sparse networks that rely on random connectivity initialization. Finally, RFs are created before training and can be modified to capture the most essential characteristics of a dataset. These attributes lend biological inspiration to our dANN models and are expected to be advantageous for neuromorphic hardware implementation, particularly when faced with space limitations and increased energy consumption resulting from lengthy node connections.

It is crucial to acknowledge the boundaries and limitations of our dANN architecture. In the implementation presented here, to maintain their initial connectivity, dANNs necessitate an extra boolean mask multiplication after every gradient descent step.

This additional step results in a higher computational expense regarding floating-point operations. Moreover, during training, we employ the backpropagation algorithm and discard some gradients that aren't linked to an existing connection, potentially losing vital information from other gradient directions that could result in faster convergence. Using locally computed gradients that are dependent only on the connected nodes would overcome these limitations and further improve the efficiency/performance of our dANN models, but such an implementation is currently not possible with existing ML platforms and requires custom-made code. Of note, the same limitations apply to sparse network architectures that are implemented using the existing ML platforms. Finally, the cable weights used here are unconstrained. A more bio-realistic approach would be to restrict these weights to positive values or set them to fixed, positive values. Nevertheless, we believe our work is valuable as it offers new insights into the benefits, i.e., improved accuracy, less overfitting, and much fewer parameters, that can be gained by adopting dendritic features in classical ANNs. These advantages combined with their ability to scale relatively well with depth, render the proposed dANN networks a potentially powerful alternative to classical ANNs through their incorporation in DL architectures such as convolutional neural nets or transformers.

Overall, we show that implementing dendritic properties can significantly enhance the learning capabilities of ANNs, making them both accurate and efficient. These findings hold great promise as they suggest that integrating biological characteristics could be crucial for optimizing the sustainability and effectiveness of ML algorithms.

## Methods

### Network architectures

We have developed a range of traditional ANNs consisting of two hidden layers and an output layer that matches the number of classes. To create the dANN model, we first create two boolean masks that determine the synaptic weights between the input and dendritic layers, as well as the cable weights between dendritic and somatic layers. Once we have initialized the model, we apply these masks to achieve a sparse network with structured synaptic and cable weights (Eq. 1).

$$W_k \leftarrow W_k \odot M_k \quad (1)$$

where $W_k$ denotes the weights, $M_k$ the boolean mask associated $k$-th layer, and $\odot$is the Hadamard (elementwise) product.

The calculation for the forward pass is obtained by linearly combining the inputs with the weights and adding the bias in each node of all layers. Finally, the output is obtained by passing the summation through a (nonlinear) activation function (Eq. 2).

$$A_k = f\left(W_k X_k + b_k\right) \quad (2)$$

where $X_k$ and $A_k$ denote the inputs to the $k$-th layer and its activations, respectively. $f(\cdot)$ is the activation function.

In the output layer, we calculate the loss, which is then propagated back to calculate the gradients with respect to all trainable parameters respectively. To ensure the same connectivity as the original model, we zero out all gradients calculated in non-existent connections (Eq. 3).

$$\frac{\partial L}{\partial \vartheta} \leftarrow \frac{\partial L}{\partial \vartheta} \odot M_k \quad (3)$$

where $L$ is the loss and $\vartheta$ denotes the trainable parameters.

**Synaptic connections**

To define the input to the dendritic connectivity matrix, we utilize three distinct strategies. Our first approach involves random allocation, where each dendrite receives 16 inputs (pixels) which are randomly selected from the image (dANN-R). Our second approach utilizes local-constructed receptive fields (dANN-LRF), where each dendrite again receives 16 inputs (pixels), but this time they are sampled from a restricted part of the image. To do so, we randomly select a pixel to represent the center of the receptive field for each dendrite. In particular, the central pixel is drawn from a uniform distribution. Then, the 16 inputs are chosen from the 4x4 neighborhood of that pixel. The process is repeated for all dendrites. Finally, we utilize a global-constructed receptive field (dANN-GRF). In this approach, we select a pixel to represent the receptive field center for each soma instead of each dendrite. Then, the central pixel of each dendrite belonging to that soma has a central pixel drawn from a uniform distribution around the central somatic pixel. Finally, dendrites receive 16 inputs from the 4x4 neighborhood of their central pixel, as before.

**Datasets**

The dANN models are trained to classify images into one of their respective classes. The MNIST[109] consists of handwritten digits from 0 to 9. Fashion MNIST[110] is an alternative to MNIST and consists of clothing images: T-shirt/top, trousers, pullover, dress, coat, sandal, shirt, sneaker, bag, and ankle boot. Kuzushiji MNIST[111] is a drop-in replacement for the MNIST dataset consisting of one Japanese character representing each of the ten rows of Hiragana. All of these datasets come with 60,000 training and 10,000 test images. Extended MNIST[112] follows the same conversion paradigm used to create the MNIST dataset. The result is a set of datasets that constitute more challenging classification tasks involving letters and digits. Here, we used the 47 balanced classes with 731,668 training and 82,587 testing images. All MNIST variants consist of 28x28 grayscale images. Finally, CIFAR-10[113] consists of images of objects or animals in ten classes: airplane, automobile, bird, cat, deer, dog, frog, horse, ship, and truck. The dataset contains 50,000 training and 10,000 test images. The images are 32x32 pixels in three color channels.

For our experiments, we trained the models with 90% of the training data, keeping the remaining 10% for validation. Once the training was complete, we evaluated the performance on the test set.

**Hyperparameters**

Our models were trained using the Adam optimization algorithm, with the default parameters of a learning rate of 0.001 and betas of 0.9 and 0.999. To ensure efficient training, we utilized a minibatch size of 128. The number of epochs was variable for each dataset but the same across models. Specifically, for MNIST, FMNIST, and KMNIST we used 15, 20, and 20 epochs, respectively, whereas for EMNIST and CIFAR10 we used 50 epochs. For the sequential learning scenario, we used 50 epochs to train the models. In our dANN models, each dendrite receives inputs from 16 input neurons. We calculate the loss using the cross-entropy function and have set the activation function of all nodes to the Leaky Rectified Linear Unit (ReLU) with a negative slope of 0.1, except for the output nodes, which utilize the softmax activation function.

**Efficiency scores**

To calculate the efficiency scores for all networks, we formulated the accuracy (*aes*) and loss (*les*) efficiency scores, respectively. To do so, we normalized the accuracy and loss with a factor $f$ that takes values from $[1,\infty]$ and is the ratio of the logarithm, with base 10, of the number of trainable parameters of a model $i$ with the minimum number of trainable parameters of the compared models times the number of epochs needed to reach minimum validation loss (Eq. 4).

$$f=\frac{\log_{10} k_i}{\min_m\left(\log_{10}\left(k_m\right)\right)} \quad (4)$$

By dividing the accuracy by the factor $f$, the score remains in $[0,100]$ while multiplying the loss with $f$ the loss score remains in $[0,\infty]$ (Eq. 5).

$$aes=\frac{accuracy(\%)}{f}, les=loss\cdot f \quad (5)$$

where $k_i$ denotes the number of trainable parameters multiplied by the number of epochs needed to reach minimum validation loss of the model $i$, and $m$ denotes the compared models.

Using these scores, models with large numbers of trainable parameters and training epochs have lower accuracy efficiency scores and higher loss efficiency scores compared to models with fewer trainable parameters and fewer training epochs.

**Interpretability analysis**

*Synaptic, cable, and output weights*: To display the learned weights, we constructed histograms with 20 bins and utilized kernel density estimation (KDE) to approximate the underlying distribution using a continuous probability density curve. The KDE plot smooths the observations with a Gaussian kernel, generating a continuous estimation. The histograms are built by concatenating the learned weights across *N=5* initializations for each ANN model.

*Entropy*: To calculate the entropy for the first and second hidden layers, we used the activations of all nodes during evaluation against the test set. First, we created the hit matrix ($HM \in R^{n_{class} \times D}$) for each layer that assigns the number of times a node was activated (activity above zero) for images belonging to the same category.

To calculate normalized probabilities, we add an extra row to *HM* containing the number of times a node remained inactive. Thus, the summation across rows is equal to the number of data samples in the test set.

Obtaining the probability matrix by dividing the *HM* by the number of images, we calculate the entropy for each node in the $k$-th layer (Eq. 6).

$$H_k = -\sum_{i=1}^{D} p(x) \log_2 p(x) \quad (6)$$

We plot the entropy distributions using histograms with 20 bins and the KDE method to estimate a continuous probability density function. We removed inactive nodes from the analysis.

*Selectivity*: To calculate how selective a node is, we calculated how many categories it was activated for. We consider activity significant if a node was activated for over 400 images of a specific category. Thus, we ended with integers in $[1, n_{class}]$, with 1 denoting class-specificity and $n_{class}$ total mixed-selectivity. As selectivity is a discrete metric, we plot the histograms with $n_{class}$ bins and without using KDE.

**Dimensionality reduction and analysis**

To calculate the representations of the hidden layers, we used the t-distributed stochastic neighbor embedding (TSNE) dimensionality reduction method[74] with perplexity equal to 50. We chose this technique based on its widespread popularity and proven ability to preserve neighborhoods and clusters in projections. Activations for a given layer, the subject of our analysis, are extracted strictly for a random subset of 2,000 observations from the test sets to aid visual presentation. We visualize projections as scatterplots, with points colored to show their class. To assess the quality of the projection and its discriminatory power, we employ two metrics. The Silhouette score calculates the global structure of the projection. It shows if activations of images belonging to the same category are close in the reduced space[114], and the neighborhood hit score (*NH*) shows the local structure in the projection and indicates how well classes are visually separated[115,116].

Silhouette score is calculated using the mean intra-cluster distance (a) and the mean nearest-cluster distance (b) for each sample (Eq. 7).

$$s_i = \frac{b - a}{\max(a, b)}, i \in \{1, 2, \ldots, n\} \quad (7)$$

and takes values in $[-1,1]$, with values closer to 1 denoting good clustering and values close to 0 indicating overlapping clusters. Negative values indicate a sample has been assigned to the wrong cluster, as a different cluster is more similar. The Silhouette score is the average over all samples.

The *NH* denotes how similar a point is to its $k$ nearest neighbors in the reduced space. For a given $k$, the *NH* for a point $x$ is the percentage of the $k$-nearest neighbors that belong to the same class as $x$ (Eq. 8).

$$NH(k)=\frac{1}{n}\sum_{i=1}^{n}\left(\frac{\sum_{j=1}^{k} l_j}{k}\right), l_j=1\, if\, r_j \in C^{[x_i]}\, else\, 0 \quad (8)$$

where $C^{[x_i]}$ denotes the category of the data point $x_i$, and $r_j$ are the point in its neighborhood. Then the neighborhood hit score is calculated as an average across all points $x$ in the dataset. The score is bounded in [0, 1], with higher values denoting better neighborhood compactness and small values of misplaced data points. Here, we use $k=11$.

Trustworthiness is a metric that expresses the extent to which the local structure is retained after the dimensionality reduction[117] (Eq. 9).

$$T(k)=1-\frac{2}{nk(2n-3k-1)}\sum_{i=1}^{n}\sum_{j\in N_i^k} max(0,(r(i,j)-k)) \quad (9)$$

where $r(i,j)$ denotes the rank of the datapoint $j$ according to the pairwise distances between the low-dimensional datapoints, and $N_i^k$ represents the $k$-nearest neighbors of datapoint $i$ in the low-dimensional space, but not in the high-dimensional space.

Thus, any unexpected nearest neighbors in the low-dimensional space are penalized proportionally to their rank in the high-dimensional space. The trustworthiness is within $[0,1]$. Here, we use $k = 11$.

**Computing resources and software**

All simulations were performed on a custom machine under the Debian GNU/Linux trixie/sid (kernel version 6.6.15-2) operating system with 64GB of RAM, Intel® Core™ i5-10400 CPU @ 2.90GHz, and an NVIDIA GeForce RTX 3080 Ti GPU @ 12GB. We implemented all models using the Keras 2.15.0 functional API[118] with TensorFlow 2.15.0 backend[119] under Python 3.9.18 (conda 23.7.4). For better handling of the training process, we used a custom training loop. For data analysis and visualization, we utilized various Python modules, including numpy 1.24.4[120], scikit-learn 1.4.1[121] , pandas 1.5.3[122], matplotlib 3.8.3[123], seaborn 0.13.2[124] and seaborn-image 0.8.0[125].

**Statistical Analysis**

For all standard statistical tests (detailed in figure legends), the significance level $\alpha$ was 0.05. To correct for multiple comparisons, $\alpha$ was divided by the number of tests according to the Bonferroni procedure. Throughout the figures, $p$ values are denoted by * ($p<0.05$), ** ($p<0.01$), and *** ($p< 0.001$). To compare the dependent value among different groups (models x layers), we used a two-way analysis of variance (ANOVA) followed by an unpaired t-test (two-tailed) with Bonferroni's correction whenever statistical difference was observed for post hoc comparisons. The statistical analysis was performed using the pingouin 0.5.4 library[126].

## Acknowledgments

The authors thank Michalis Pagkalos, Dr. Athanasia Papoutsi, Prof. Blake Richards, Ioannis-Rafail Tzonevrakis, Dr. Eirini Troullinou, and Prof. Grigorios Tsagkatakis for their valuable and constructive feedback. We also thank the reviewers for their detailed and highly constructive feedback. This work was supported by NIH (1R01MH124867-04) to P.P., the European Commission (H2020-FETOPEN-2018-2019-2020-01, FET-Open Challenging Current Thinking, NEUREKA GA-863245) to P.P.

## Author contributions

S.C. and P.P. conceived the project, designed the experiments, and wrote the manuscript. S.C. implemented the models, performed the experiments, analyzed the data, and created the figures. P.P. supervised and funded the project.

## Competing interests

The authors declare no competing interests.

## Data availability

The source code that generates all Figures and the data that support this study are accessible on GitHub.

## Computer code

The code underlying this study is available on GitHub.

## Supplementary Material

**Supplementary Table 1.** Minimum test loss scores obtained by each model on five benchmark datasets across various configurations and their corresponding test accuracy. Performance accuracy is listed as mean ± standard deviation across *N=5* initializations for each model.

| | **MODEL PERFORMANCE – MINIMUM LOSS** | | | | |
|---|---|---|---|---|---|
| **Models** | **MNIST** | **FMNIST** | **KMNIST** | **EMNIST** | **CIFAR10** |
| | **Test loss** | | | | |
| dANN-R | 0.0644±0.0013 | 0.3152±0.0093 | 0.4880±0.0044 | 0.6282±0.0068 | 1.3771±0.0067 |
| dANN-LRF | 0.0483±0.0018 | **0.2871±0.0021** | **0.4342±0.0054** | 0.5675±0.0070 | 1.2528±0.0141 |
| dANN-GRF | **0.0471±0.0022** | 0.2879±0.0040 | 0.4615±0.0133 | 0.5317±0.0034 | 1.2732±0.0065 |
| dANN-F | 0.0498±0.0020 | 0.2909±0.0067 | 0.4356±0.0104 | **0.5210±0.0018** | **1.2133±0.0087** |
| vANN | 0.0864±0.0089 | 0.3579±0.0072 | 0.6905±0.0255 | 0.6861±0.0101 | 1.5297±0.0256 |
| | **Test accuracy (%)** | | | | |
| dANN-R | 98.090±0.0583 | 89.108±0.1955 | 87.260±0.1346 | 80.986±0.2621 | 52.044±0.0172 |
| dANN-LRF | 98.466±0.1058 | **90.194±0.0543** | 89.070±0.1144 | 82.638±0.2798 | 56.076±0.3200 |
| dANN-GRF | **98.576±0.0809** | 89.996±0.2132 | 88.460±0.1872 | 83.520±0.1262 | 55.478±0.2848 |
| dANN-F | 98.442±0.0765 | 89.984±0.1227 | 89.128±0.1979 | **83.680±0.1944** | **58.176±0.4343** |
| vANN | 97.858±0.2811 | 88.122±0.1492 | **91.194±0.3330** | 80.219±0.2291 | 46.680±0.8106 |

**Supplementary Table 2.** Characterization of probability density functions is shown in Figure 4**A**. Skewness, kurtosis, and range are calculated.

| **models** | **First hidden layer** | | | **Second hidden layer** | | | **Output layer** | | |
|---|---|---|---|---|---|---|---|---|---|
| | **kurtosis** | **skewness** | **range** | **kurtosis** | **skewness** | **range** | **kurtosis** | **skewness** | **range** |
| dANN-R | 2.741 | -0.308 | 2.961 | 0.088 | -0.124 | 3.261 | 0.857 | -0.753 | 3.092 |
| dANN-LRF | 2.157 | -0.372 | 4.082 | -0.103 | 0.074 | 4.540 | 1.192 | -0.510 | 3.496 |
| dANN-GRF | 3.457 | -0.545 | 4.700 | 0.638 | 0.001 | 4.173 | 1.393 | -0.561 | 2.869 |
| vANN | 2.814 | -0.234 | 2.019 | 2.059 | 0.090 | 1.373 | 1.905 | -0.796 | 0.805 |

**Supplementary Table 3.** Test accuracy obtained by each model on five noise levels, i.e., increasing the standard deviation of the Gaussian noise, and their corresponding test loss against the FMNIST noisy dataset. Test accuracy and loss are listed as mean ± standard deviation across *N=5* initializations for each model.

| | **MODEL PERFORMANCE – NOISY IMAGES** | | | | |
|---|---|---|---|---|---|
| **Models** | **σ=0.0** | **σ=0.25** | **σ=0.5** | **σ=0.75** | **σ=1.0** |
| | **Test accuracy (%)** | | | | |
| dANN-R | 89.392±0.2071 | 83.388±0.2340 | 77.108±0.2064 | 70.206±0.3007 | 64.456±0.5200 |
| dANN-LRF | 89.310±0.0514 | 83.926±0.2560 | **77.744±0.1853** | **71.73±0.1393** | **66.288±0.3290** |
| dANN-GRF | **89.616±0.3628** | 83.718±0.1289 | 76.864±0.3725 | 70.490±0.4438 | 64.996±0.2247 |
| dANN-F | **89.520±0.1814** | 83.766±0.2763 | 77.392±0.4827 | 71.386±0.6403 | 65.942±0.1788 |
| vANN | 89.288±0.3654 | **84.178±0.2922** | 77.054±0.6884 | 70.624±0.5920 | 64.186±1.2748 |
| | **Test loss** | | | | |
| dANN-R | 0.3167±0.0035 | 0.5896±0.0078 | 0.8985±0.0072 | 1.176±0.0221 | 1.3742±0.0206 |
| dANN-LRF | 0.2997±0.0018 | **0.4519±0.0081** | **0.6224±0.0079** | **0.7826±0.0098** | **0.9251±0.0045** |
| dANN-GRF | **0.2932±0.0086** | 0.4868±0.0113 | 0.7033±0.0159 | 0.8827±0.0202 | 1.0381±0.0185 |
| dANN-F | 0.2955±0.0038 | 0.4658±0.0111 | 0.6491±0.0156 | 0.8124±0.0119 | 0.9560±0.0086 |
| vANN | 0.404±0.0066 | 0.9869±0.0492 | 1.6652±0.0528 | 2.196±0.0583 | 2.6139±0.0827 |

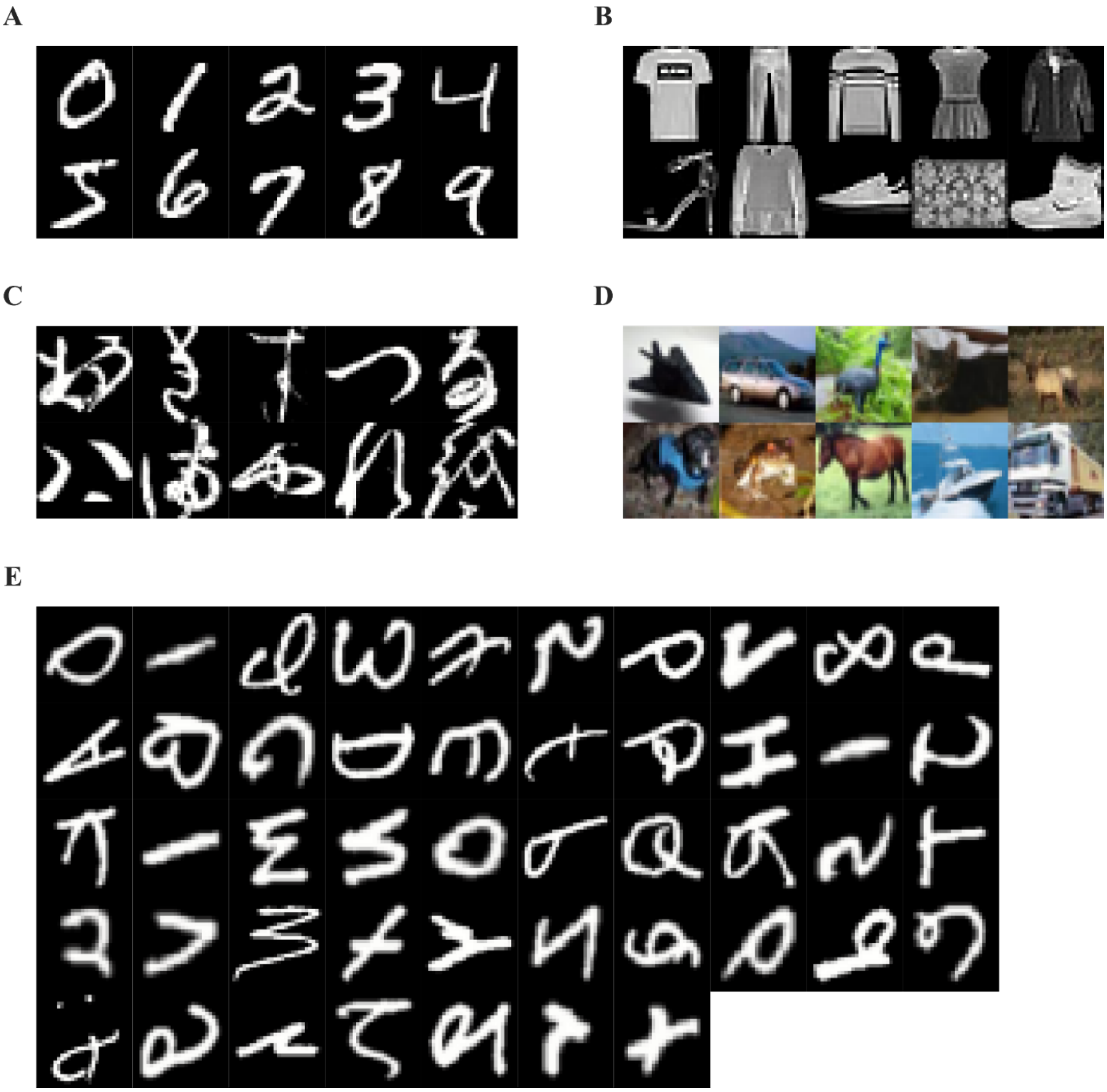

**Supplementary Figure 1. The classification benchmark datasets. A.** The MNIST dataset. **B.** The Fashion MNIST dataset. **C.** The Kuzushiji MNIST dataset. **D.** The CIFAR10 dataset. **E.** The EMNIST dataset.

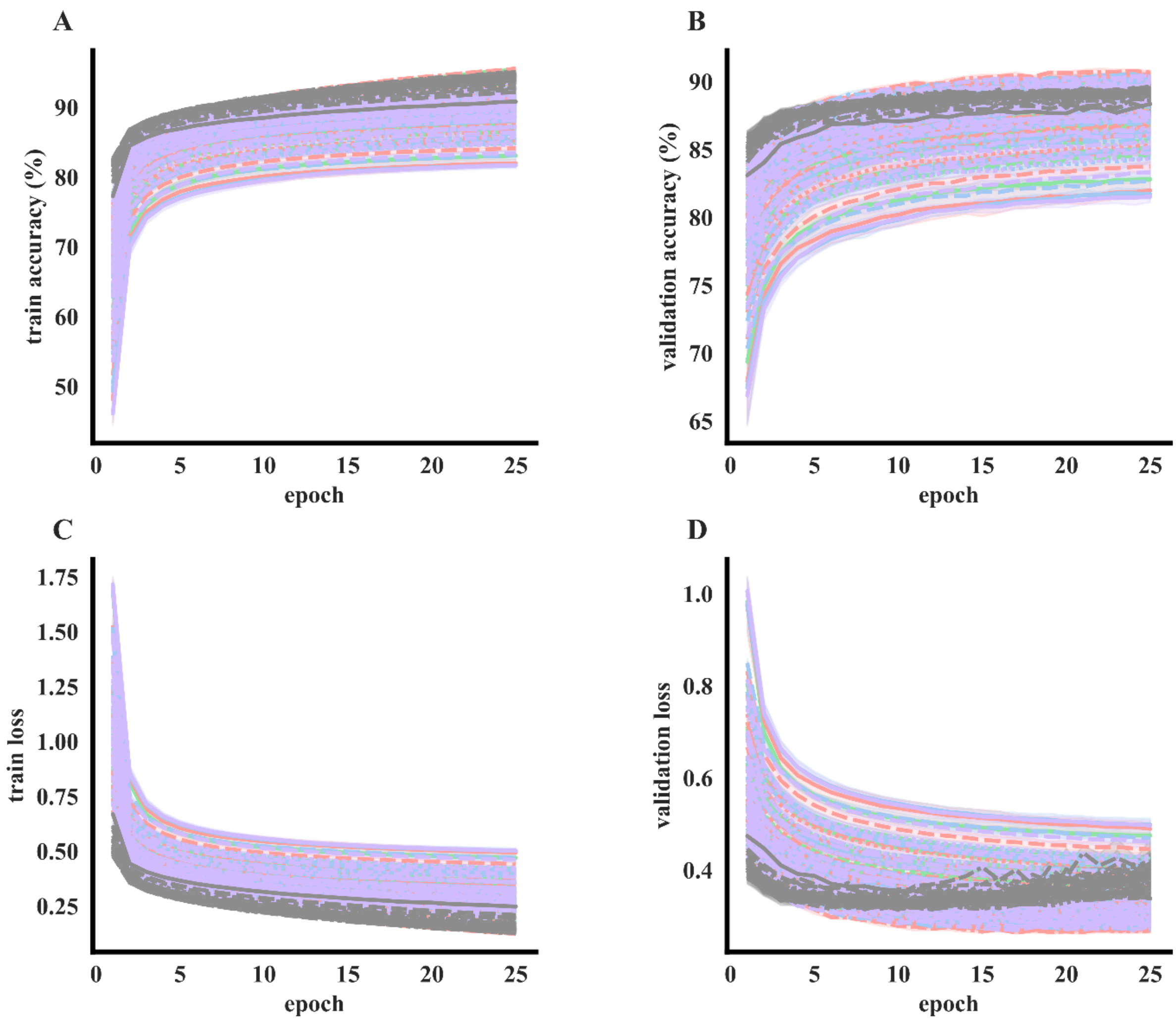

**Supplementary Figure 2. Training and validation process of all models on FMNIST. A.** Training accuracy for all model configurations (i.e., number of nodes in the hidden layers). **B.** Validation accuracy. **C.** Train loss. **D.** Validation loss. Shades denote standard deviation calculated across *N=5* initializations for each model.

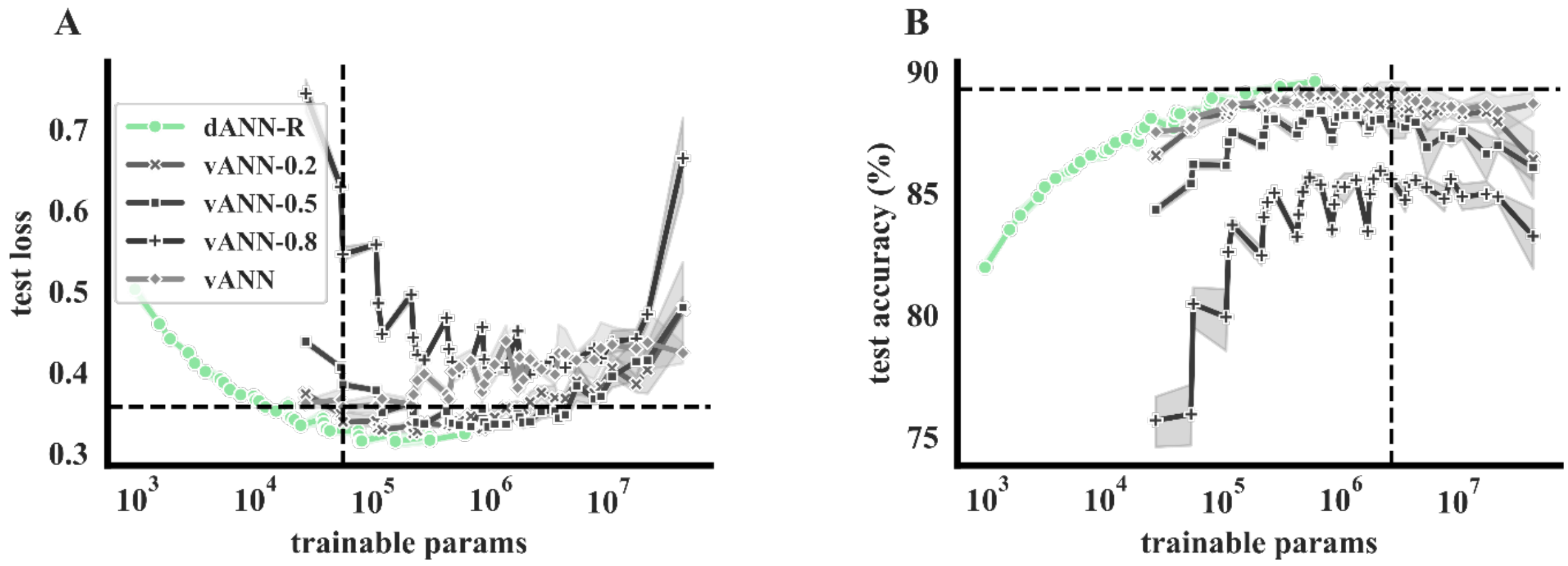

**Supplementary Figure 3. dANNs vs vANNs with added dropout layers on FMNIST. A.** Average test loss as a function of the trainable parameters of the models used: A dendritic ANN (dANN-R) with random inputs (green), and four variants of vANN (grey shades), with 0.2, 0.5, 0.8 dropout rates or no dropout layers, respectively. Horizontal and vertical dashed lines denote the minimum test loss of the vANN and its trainable parameters, respectively. The x-axis is shown in a logarithmic scale (log10). **B.** Similar to **A**, but depicting the test accuracy instead of the loss.

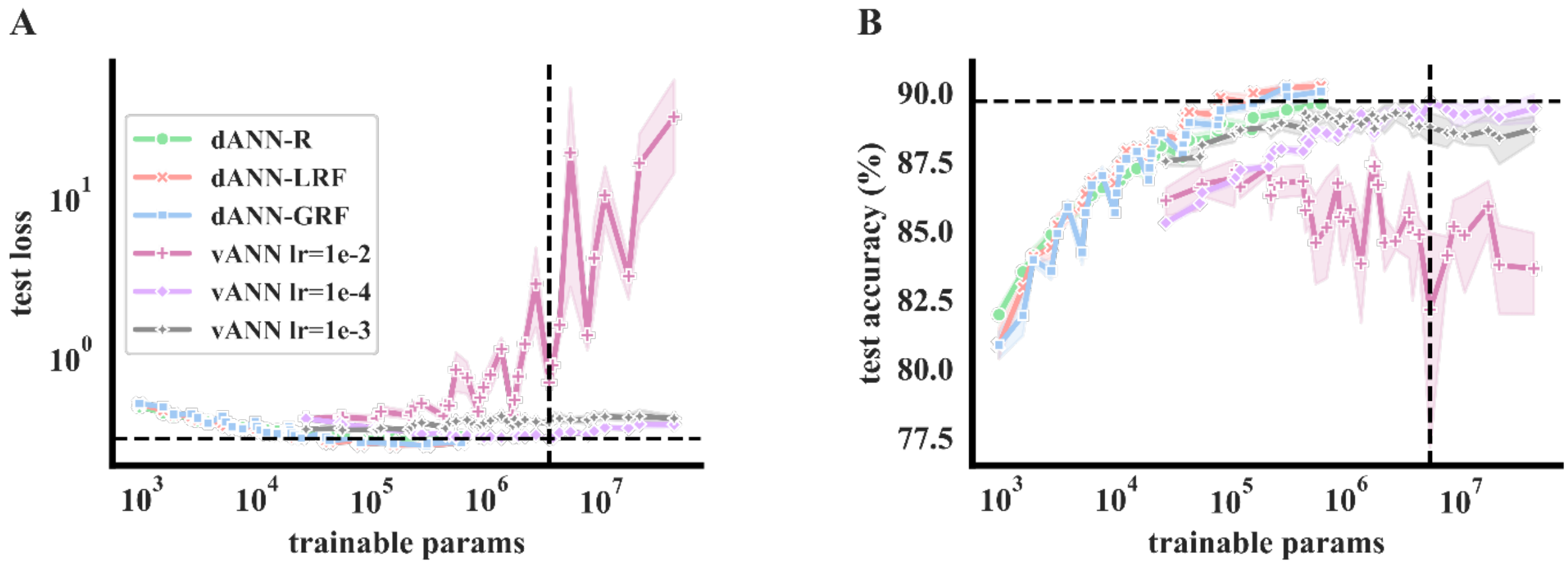

**Supplementary Figure 4. Effect of learning rate on vANNs for the FMNIST dataset. A.** Average test loss as a function of the number of trainable parameters of the six models used: Three dendritic ANNs with Random, LRF and GRF input sampling, and three vANN models: with 1e-2 (dark pink), 1e-3 (grey), and 1e-4 (light purple) learning rates, respectively. Horizontal and vertical dashed lines denote the minimum test loss of the control vANN and its trainable parameters, respectively. Both axes are shown in a logarithmic scale (log10). **B.** Similar to **A**, but depicting the maximum test accuracy instead of the loss. Note that the y-axis is shown in a linear scale here.

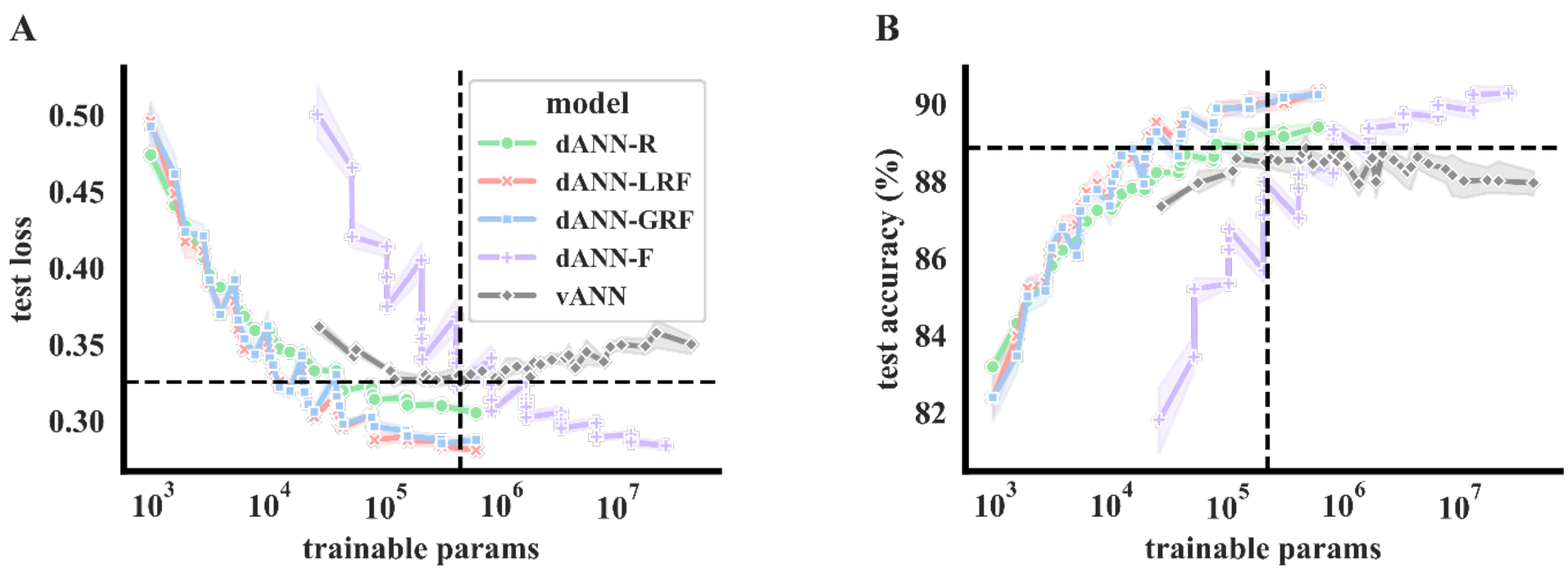

**Supplementary Figure 5. Effect of early stopping on model performance on FMNIST. A.** Average test loss as a function of the trainable parameters of the five models used: A dendritic ANN (dANN-R) with random inputs (green), a dANN with LRFs (red), a dANN with GRFs (blue), a dANN (dANN-F) with all-to-all inputs (orange), and the vANN model (grey). Horizontal and vertical dashed lines denote the minimum test loss of the vANN and its trainable parameters, respectively. The x-axis is shown in a logarithmic scale (log10). **B.** Similar to **A**, but depicting the test accuracy instead of the loss.

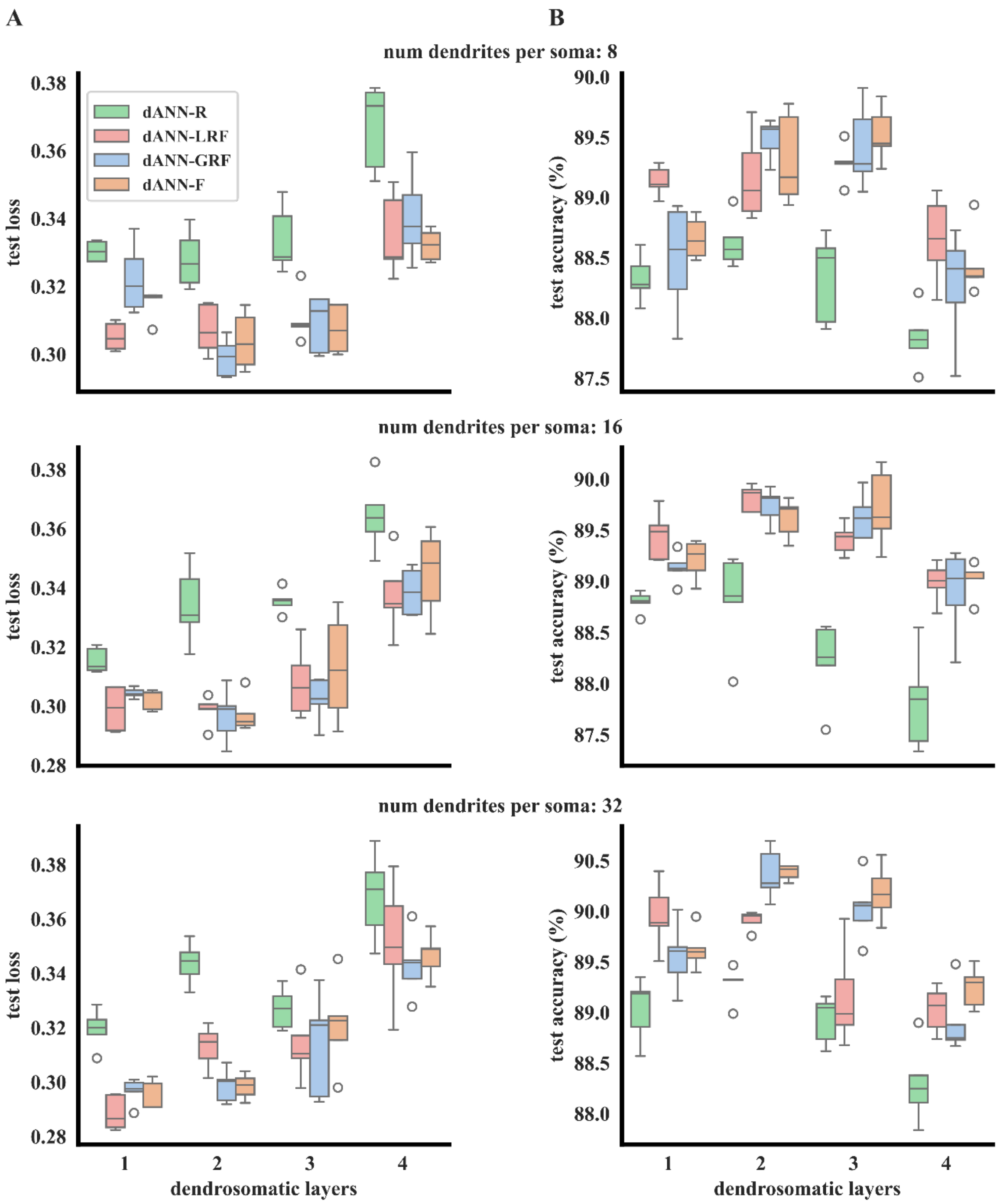

**Supplementary Figure 6. dANNs trained on FMNIST can scale with depth. A.** Test loss as a function of the number of dendrosomatic layers for the four classes of dANNs used: A dendritic ANN (dANN-R) with random input sampling (green), a dANN with LRFs (red), a dANN with GRFs (blue), and a dANN with all-to-all input sampling (orange). Horizontal and vertical dashed lines denote the minimum test loss of the vANN and its trainable parameters, respectively. The x-axis is shown in a logarithmic scale (log10). **B.** Similar to **A**, but depicting the test accuracy instead of the loss. Top to bottom: increasing number of dendrites per soma: 8, 16, 32, respectively. Number of somata was held constant to 256.

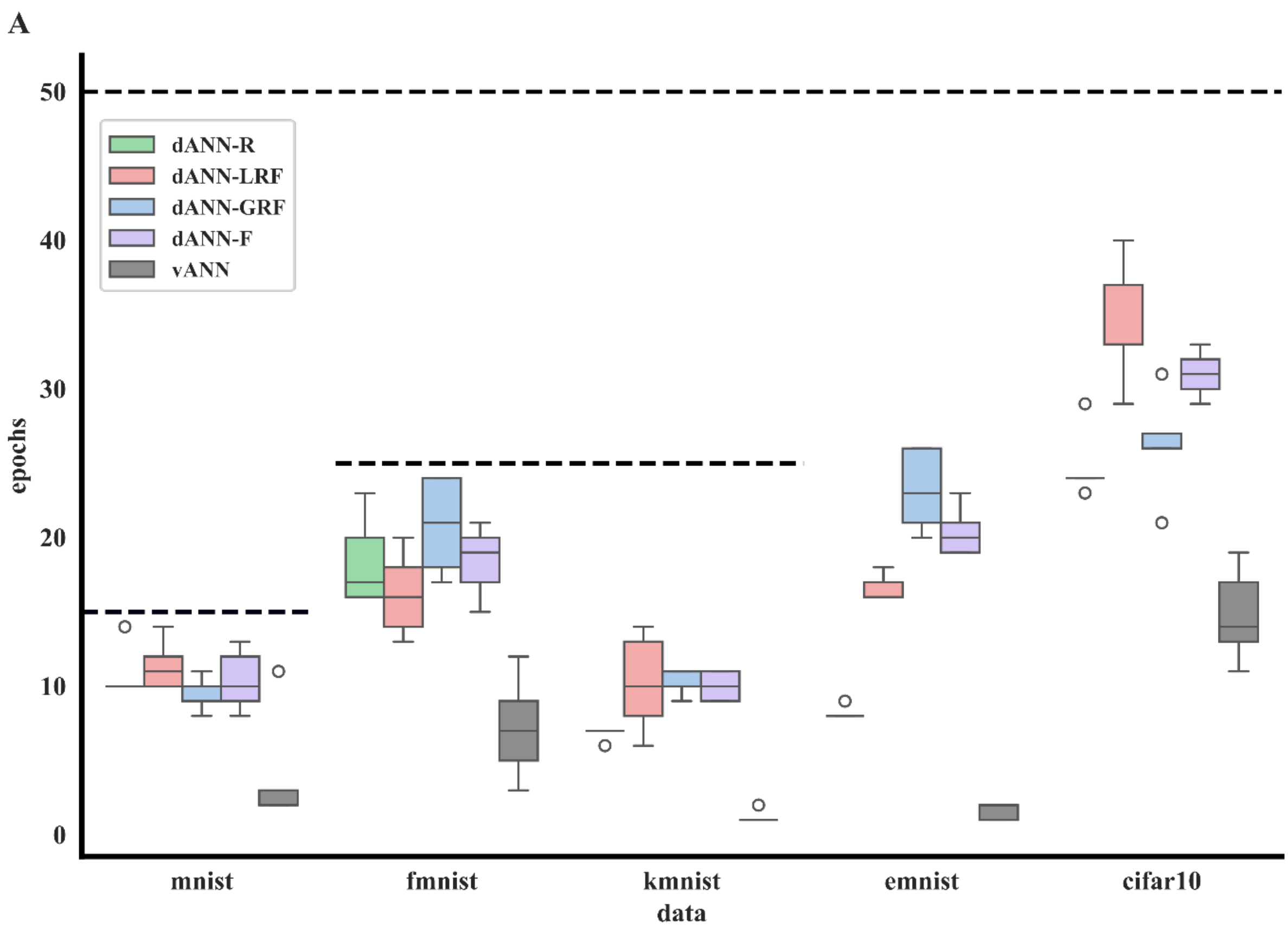

**Supplementary Figure 7. Number of epochs needed to reach minimum validation loss. A.** Number of epochs where validation loss was minimum for the dANN-R (green), dANN-LRF (red), dANN-GRF (blue), dANN-F (purple) and vANN (grey). Dashed lines indicate the number of epochs used for training: for MNIST, 15 epochs were used. For FMNIST and KMNIST, 25 epochs were used. For EMNIST and CIFAR10, 50 epochs were used.

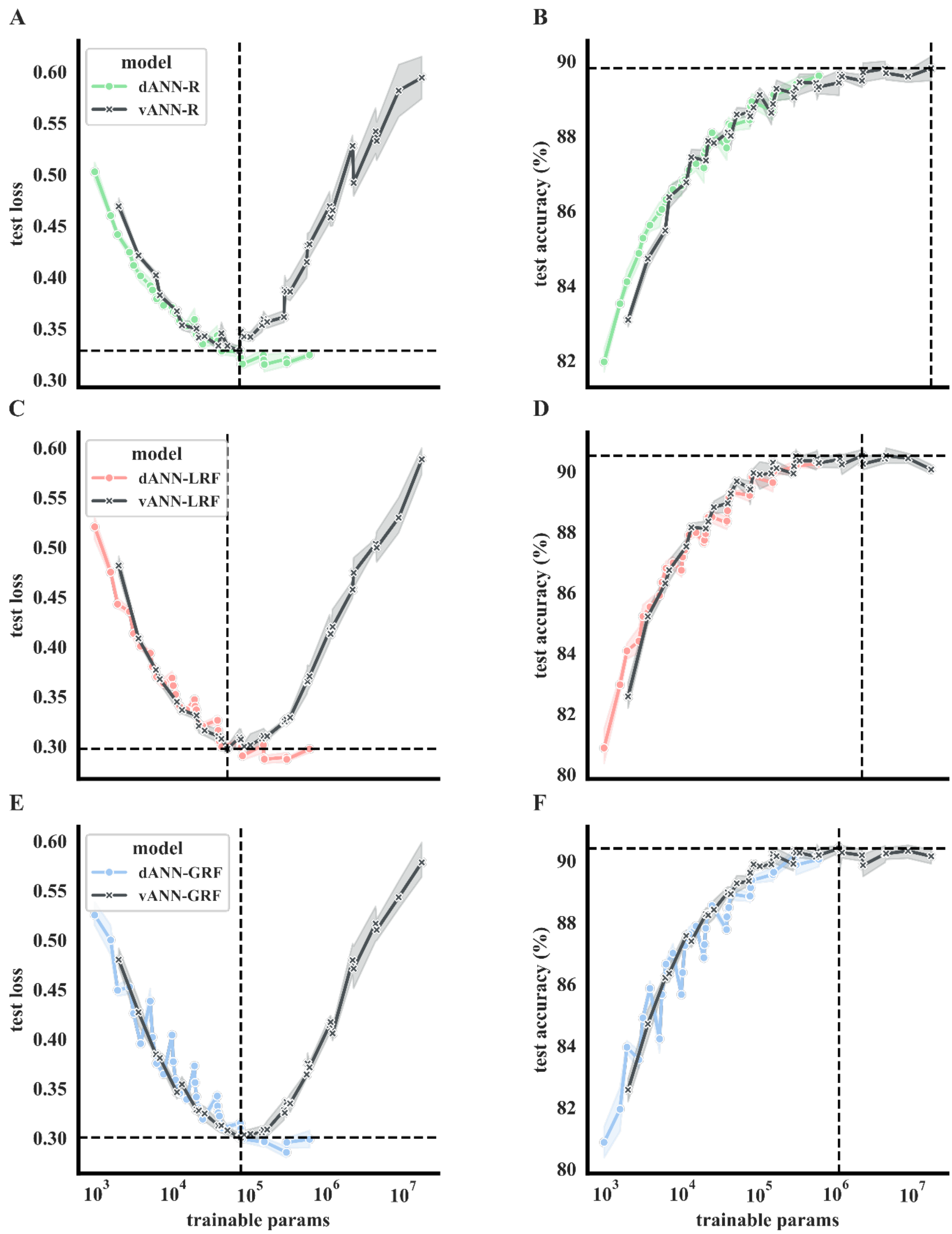

**Supplementary Figure 8. Training and validation process of all models. A.** Average test loss as a function of the trainable parameters of the dANN-R and vANN-R models. Horizontal and vertical dashed lines denote the minimum test loss of the vANN-R and its trainable parameters, respectively. The x-axis is shown in a logarithmic scale (log10). **B.** Similar to B, but depicting the test accuracy instead of the loss.

From top to bottom: random inputs, local receptive fields, global receptive fields. **C-D.** Similar to **A** and **B**, respectively, but depicting here the dANN-LRF and vANN-LRF. **E-F.** Similar to **A** and **B**, respectively, but depicting here the dANN-GRF and vANN-GRF.

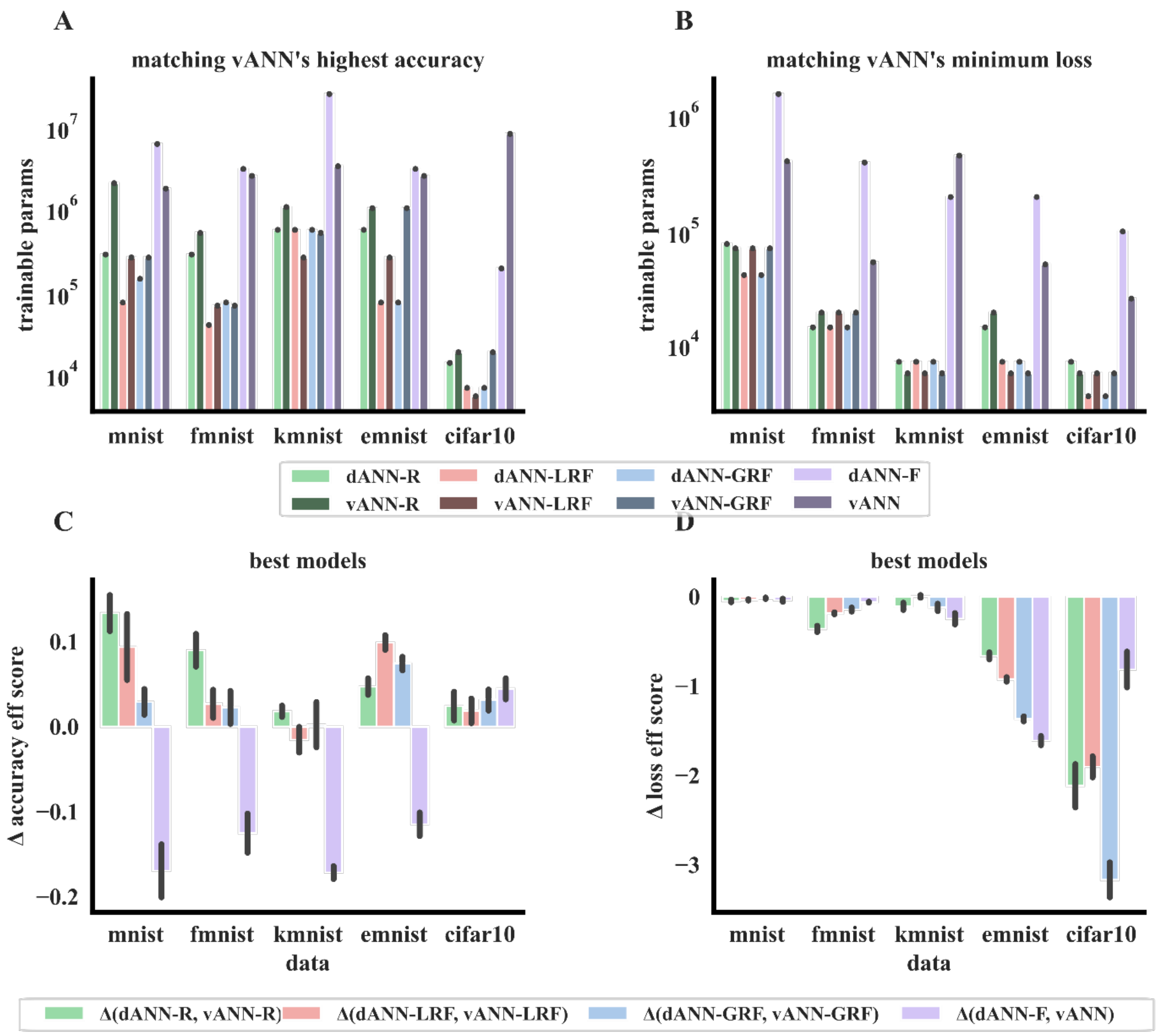

**Supplementary Figure 9. dendritic vs. vANNs with sparse sampling. A.** Number of trainable parameters that each model needs, dANN and vANN-R with random input sampling (light and dark green), dANN-LRF and vANN-LRF (light and dark red), dANN-GRF and vANN-GRF (light and dark blue), and dANN-F and vANN (light and dark purple) to match the highest test accuracy of the respective vANN. **B.** The same as in **A**, but showing the number of trainable parameters required to match the minimum test loss of the vANN. **C.** Difference (Δ) in accuracy efficiency score for dANN - vANN with all input sampling methods and all datasets tested. Test accuracy is normalized with the logarithm of trainable parameters times the number of epochs needed for convergence. The score is bounded in [0, 1]. **D.** Same as in **C**, but showing the difference (Δ) loss efficiency score. Again, we normalized the test score with the logarithm of the trainable parameters times the number of epochs needed for convergence. The score is bounded in [0, ∞). In all barplots the errorbars represent three standard deviations across N=5 initializations for each model.

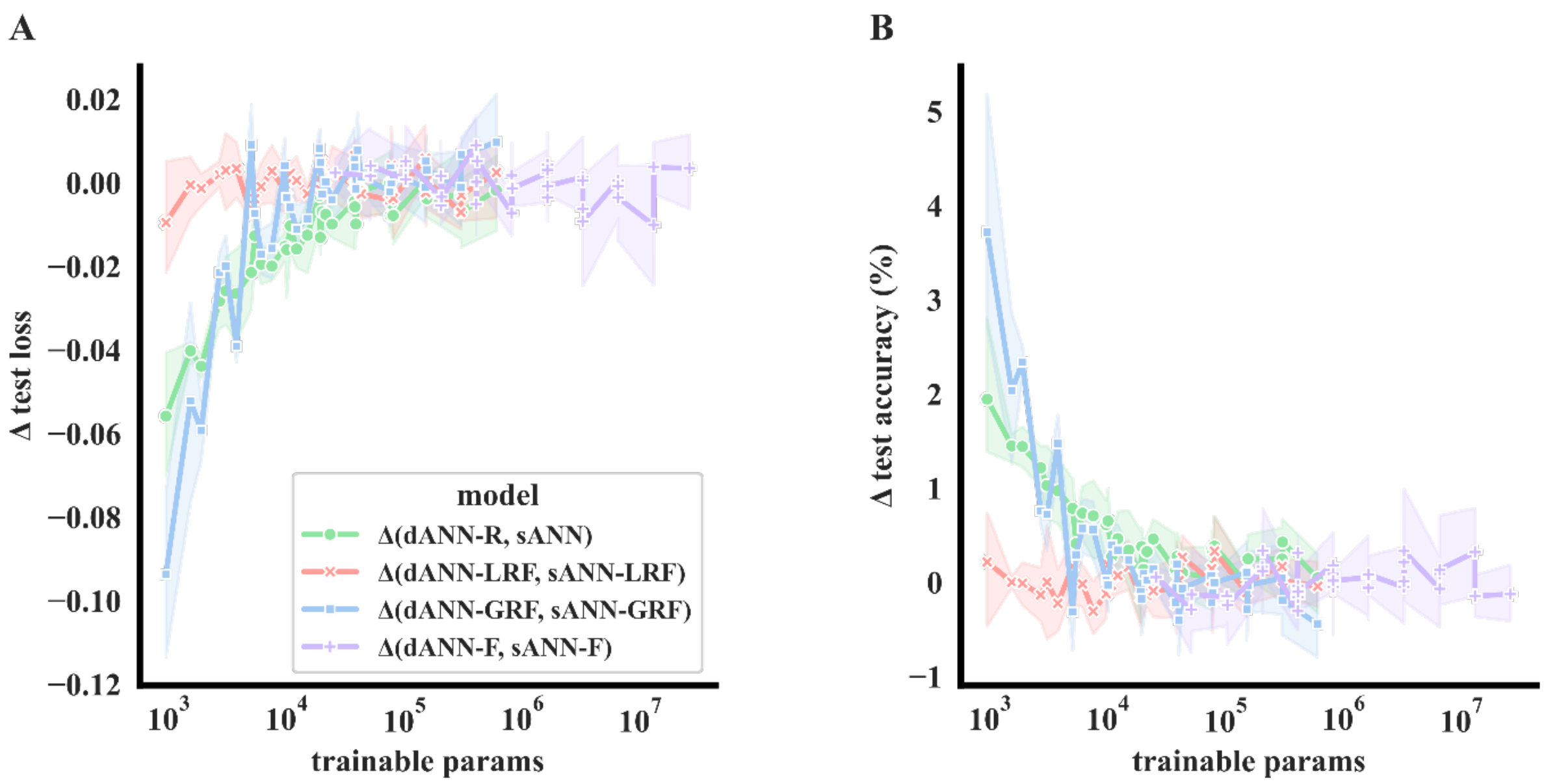

**Supplementary Figure 10. Dendritic vs. random sparse ANNs. A.** Average difference of test loss subtracting the sANN's loss from the dANN's one using models with the same input sampling method: random (green), LRF (red), GRF (blue), and all-to-all (orange). **B.** Similar to **A**, but depicting the difference in test accuracy instead of the test loss.

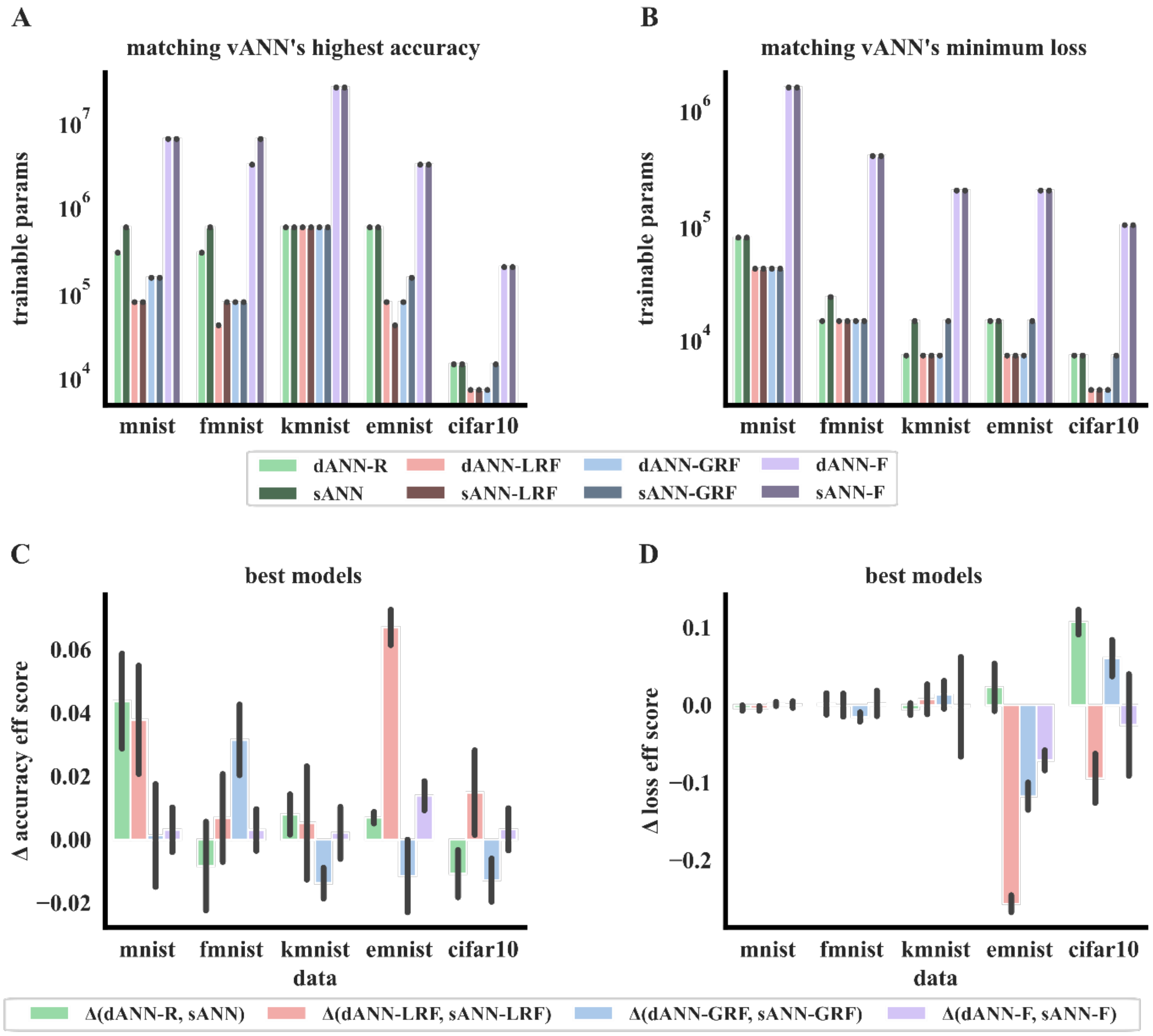

**Supplementary Figure 11. dendritic vs. sparse ANNs. A.** Number of trainable parameters that each model needs to match the highest accuracy of the vANN. Colors: dANN and sANN with random input sampling (light and dark green), dANN-LRF and sANN-LRF (light and dark red), dANN-GRF and sANN-GRF (light and dark blue), and dANN-F and sANN-F (light and dark purple) to match the highest test accuracy of the respective vANN. **B.** The same as in **A**, but showing the number of trainable parameters required to match the minimum test loss of the vANN. **C.** Difference (Δ) in accuracy efficiency score for the dANN – the sANN for all input sampling methods and all datasets tested. Test accuracy is normalized with the logarithm of trainable parameters times the number of epochs needed for convergence. The score is bounded in [0, 1]. **D.** Same as in **C**, but showing the difference (Δ) of the loss efficiency score. Again, we normalized the test score with the logarithm of the trainable parameters times the number of epochs

needed for convergence. The score is bounded in [0, ∞). In all barplots the errorbars represent three standard deviations across N=5 initializations for each model.